\documentclass{article}

\usepackage{microtype}
\usepackage{graphicx}
\usepackage{subfig}
\usepackage{booktabs}
\usepackage{todonotes}
\usepackage{enumitem}

\usepackage{hyperref}

\usepackage[accepted]{icml2025}

\usepackage{amsmath}
\DeclareMathOperator*{\argmax}{arg\,max}
\usepackage{amssymb}
\usepackage{mathtools}
\usepackage{amsthm}

\usepackage[capitalize,noabbrev]{cleveref}

\theoremstyle{plain}

\theoremstyle{definition}

\theoremstyle{remark}

\usepackage{xcolor}
\usepackage{lipsum}

\icmltitlerunning{Loss Landscape Analysis for Reliable Quantized ML Models for Scientific Sensing}

\begin{document}

\twocolumn[
\icmltitle{Loss Landscape Analysis for Reliable Quantized ML Models \\for Scientific Sensing}





\icmlsetsymbol{equal}{*}

\begin{icmlauthorlist}
\icmlauthor{Tommaso Baldi}{SSSA}
\icmlauthor{Javier Campos}{FNAL}
\icmlauthor{Olivia Weng}{UCSD}
\icmlauthor{Caleb Geniesse}{LBNL}
\icmlauthor{Nhan Tran}{FNAL}
\icmlauthor{Ryan Kastner}{UCSD}
\icmlauthor{Alessandro Biondi}{SSSA}
\end{icmlauthorlist}
\vskip 0.2in
\centering

$^1$Department of Excellence in Robotics and AI, Scuola Superiore Sant’Anna, Pisa, Italy

$^2$Fermi National Accelerator Laboratory, Batavia, IL, USA

$^3$University of California San Diego, San Diego, CA, USA

$^4$Lawrence Berkeley National Laboratory, Berkeley, CA, USA
\icmlaffiliation{SSSA}{Department of Excellence in Robotics and AI, Scuola Superiore Sant’Anna, Pisa, Italy}
\icmlaffiliation{FNAL}{Fermi National Accelerator Laboratory, Batavia, IL, USA}
\icmlaffiliation{UCSD}{University of California San Diego, San Diego, CA, USA}
\icmlaffiliation{LBNL}{Lawrence Berkeley National Laboratory, Berkeley, CA, USA}

\icmlcorrespondingauthor{Tommaso Baldi}{tommaso.baldi@santannapisa.it}

\icmlkeywords{Quantization, Loss landscape, Science ML, Robustness}

\vskip 0.3in
]




\begin{abstract}

In this paper, we propose a method to perform empirical analysis of the loss landscape of machine learning (ML) models. The method is applied to two ML models for scientific sensing, which necessitates quantization to be deployed and are subject to noise and perturbations due to experimental conditions. 
Our method allows assessing the robustness 
of ML models to such effects as a function of quantization precision and under different regularization techniques---two crucial concerns that remained underexplored so far.
By investigating the interplay between performance, efficiency, and robustness by means of loss landscape analysis, we both established a strong correlation between gently-shaped landscapes and robustness to input and weight perturbations and observed other intriguing and non-obvious phenomena. Our method allows a systematic exploration of such trade-offs \textit{a priori}, i.e., without training and testing multiple models, leading to more efficient development workflows. This work also highlights the importance of incorporating robustness into the Pareto optimization of ML models, enabling more reliable and adaptive scientific sensing systems.
\end{abstract}

\section{Introduction}
\label{sec:introduction}


Advances in sensing technology drive the frontiers of scientific exploration, enabling breakthroughs across disciplines.
Scientific sensing challenges can far outpace industrial applications, with inference latency on the scale of nanoseconds and microseconds, and extreme data rates~\cite{Duarte:2022hdp, Wei:2023mma}.  This requires new methodologies for ultra-fast and ultra-compact edge processing.

Machine learning (ML) has emerged as a transformative tool across several scientific domains. In low-latency applications, scientists can enhance the capabilities of their instruments, including adaptiveness to dynamic conditions and the extraction of deeper insights from raw data~\cite{Deiana:2021niw}.  Sensing and control with ML at unprecedented spatial and temporal scales can enable real-time analytics in scientific systems to accelerate scientific discovery.  

Although significant advances have been made in methods to co-design efficient ML algorithms to meet the performance and resource demands of scientific systems, such as quantization~\cite{survey_quant, survey_quant2}, pruning~\cite{pruning_survey_2, pruning_survey}, and neural architecture search~\cite{nas_survey}, the link between \emph{reliability} and, more in general, \emph{robustness} of models and their performance and resource optimization has been much less studied.   
However, this issue is crucial for the unique demands of scientific sensing, in which raw and unprocessed instrument data are typically exposed to harsh environments that make ML processing prone to noise and perturbations.  

\begin{figure}[tbh!]
        \centering
        \subfloat[]{%
            \includegraphics[width=0.6\linewidth]{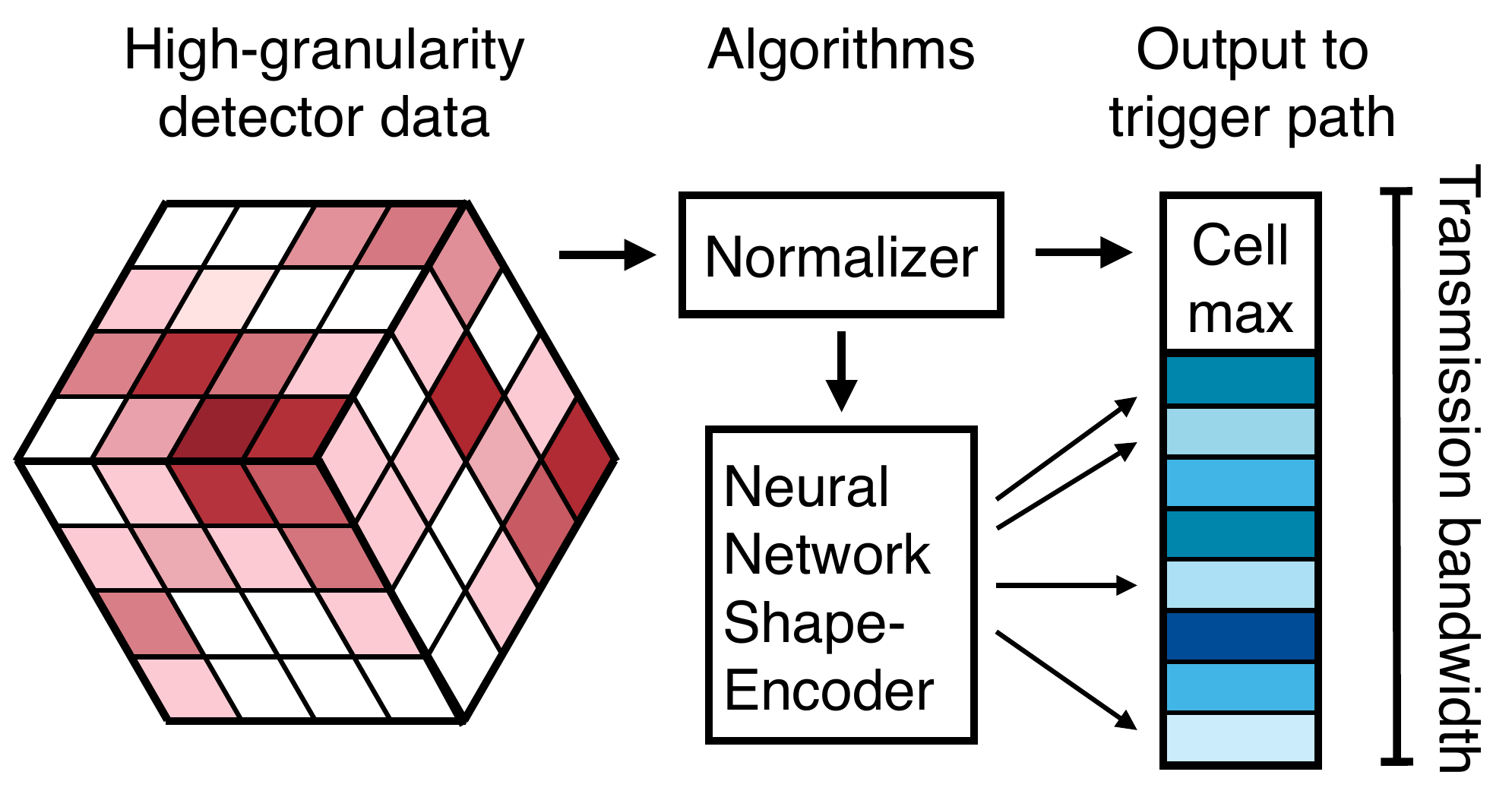}%
            \label{subfig:econ-fig}%
        }\\
        \subfloat[]{%
            \includegraphics[width=0.9\linewidth]{figs/fusion_flow.pdf}%
            \label{subfig:fision-fig}%
        }
        \caption{Workflow of the models in this study. (a) The ECON-T model workflow \cite{econ}, demonstrating the lossy data compression pipeline designed for deployment in the high-radiation environment of the Large Hadron Collider (LHC). (b) The Fusion model workflow \cite{Wei:2023mma}, illustrating active feedback control in magnetic confinement fusion devices.}
        \label{fig:models}
\end{figure}

\textbf{Contribution.} This paper addresses the interplay between performance, efficiency, and robustness in scientific sensing. Specifically, we explore the robustness of state-of-the-art (SoTA) neural networks under quantization and data corruption, focusing on two distinct applications: \textbf{(i)} autoencoders for sensor lossy data compression in particle physics \cite{econ} and \textbf{(ii)} computer vision regression tasks for fusion energy diagnostics \cite{Wei:2023mma}. An overview of their workflow is shown in Figure \ref{fig:models}, which will be later discussed in Section \ref{sec:setup}.

Specifically, the paper makes the following contributions: 
\begin{itemize}[]
    \item We introduce \emph{loss landscapes analysis}~\cite{loss_survey} methods for scientific sensing capable of identifying robust configurations of ML models \textit{a priori}, i.e., without requiring time-consuming exploration campaigns with training and testing in the loop.  
    \item We study how different regularization methods can mitigate noise and perturbations in quantized ML models for scientific sensing.
    \item We unveil a strong correlation between gently-shaped landscapes, both locally and globally, and robustness to data corruption. Furthermore, we observe non-obvious phenomena that suggest the need for a careful trade-off exploration in quantizing ML models to balance precision with robustness.
\end{itemize}

This study emphasizes the importance of including robustness to Pareto optimization of ML models, in addition to performance and efficiency, when designing real-time models for scientific sensing. By providing insights on robustness \textit{a priori}, independently of the source of noise and perturbations that may affect a model, this work paves the way for more adaptive experimental capabilities, thereby enabling more capable experiments at unprecedented timescales. 

\textbf{Paper structure.} The paper is structured as follows. We first provide basic concepts of quantization and loss landscape analysis in Section~\ref{sec:related-works}. We proceed by presenting our method based on multiple loss landscape metrics in Section~\ref{sec:methods}. Then, we introduce the setup used in this work in Section~\ref{sec:setup}, illustrating the models, the benchmarks, and the mitigation techniques involved. Experimental results are presented in Section~\ref{sec:results}, while Section~\ref{sec:conclusion} concludes the paper.

\section{Related Work}
\label{sec:related-works}
\textbf{Quantization.} Quantizing deep neural networks is a widely used technique to reduce memory usage and enhance inference speed, making it particularly valuable for deployment on resource-constrained devices. However, these benefits often come at the expense of performance degradation and increased instability. To address this issue, a popular approach is Quantization-Aware Training (QAT), where we re-train the Neural Network (NN) model with quantized parameters so that the model can recover part of the performance by converging to a better loss point. Since it is not always possible to re-train the model due to computational costs or unavailability of the dataset, an alternative approach, called Post-Training Quantization (PTQ), allows quantizing all parameters without re-training the model, with limited overhead at the cost of lower accuracy, especially for low-precision quantization \cite{ptq2020, ptq2021, ptq2022, ptq2022b}. We focus this work on QAT mainly for two reasons: first, we are interested in studying the impact of quantization on the training of NN models; and second, the training time of the target models is low enough to favor the performance advantages provided by QAT. Regardless of the quantization method, in this work, we chose uniform integer quantization due to its superior hardware efficiency compared to Floating Point (FP) representation, as evidenced by \cite{integer-only} and \cite{int_vs_fp}.

\textbf{Loss landscape Analysis.} Loss landscapes and the connections to training optimization techniques have been crucial research paths in ML for years. \cite{loss_2015, large_batch2016, loss_2019} are works on the connection between the loss landscapes and the Stochastic Gradient Descent (SGD) optimization. \cite{loss2021, loss_landscape} propose empirical analysis to better understand how several factors, such as quality of the data, number of parameters and hyperparameter tuning impact the generalization capability of the model. Our work took inspiration from the experiments of \cite{loss_landscape}, but with a different aim. As far as we know, this is the first work that looks for correlations between loss landscape topology and model robustness in science.

\section{Method}
\label{sec:methods}
This section presents the method used in this work to visualize and analyze the loss landscape of ML models. A collection of metrics is presented for analysis purposes. Our notation aligns with the conventions established by \cite{loss_landscape}.

\subsection{Loss Landscape Visualization}
This work presents plots that approximate the surface of the loss landscape, which are useful to interpret the results we obtained. Various techniques to generate this kind of plot were proposed in previous work~\cite{visulazing2014, visualizing2016, visualizing2017, visualizing2017b, visualizing}. To generate our plots, we took inspiration from the approach of~\cite{visualizing}, where the parameters of the model are perturbed along one random direction and its orthogonal, normalizing the weights filter-wise. Formally, given the parameters $\theta$ of a model, the resulting plots depict the following function:
\begin{equation}\label{eqn:loss-land}
    f(\alpha, \beta) = \mathcal{L}(\theta + \alpha\sigma + \beta\eta),
\end{equation}
where $\sigma$ and $\eta$ are the two directions and $\alpha$ and $\beta$ are the steps in these directions. A series of $N$ steps can be computed as $\alpha_i, \beta_i = \nu_{\text{min}} + i \cdot (\nu_{\text{max}} - \nu_{\text{min}})/(N - 1),$ for $i = 0, 1, \dots, N-1$, where $\nu_{\text{max}}$ and $\nu_{\text{min}}$ are the maximum and minimum perturbation module, respectively. 2D plots can be obtained by fixing either $\alpha$ or $\beta$.

\begin{figure}[ht!]
        \centering
        \subfloat[]{%
            \includegraphics[width=0.44\linewidth]{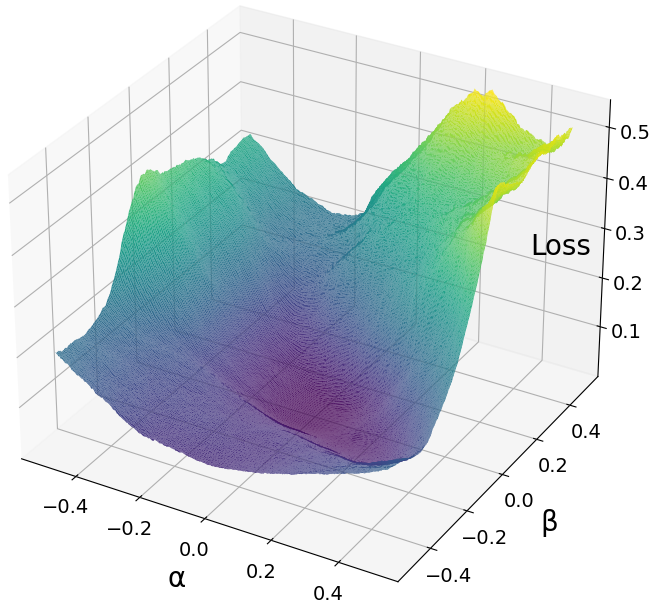}%
            \label{subfig:hessian-loss}%
        }%
        \subfloat[]{%
            \includegraphics[width=0.45\linewidth]{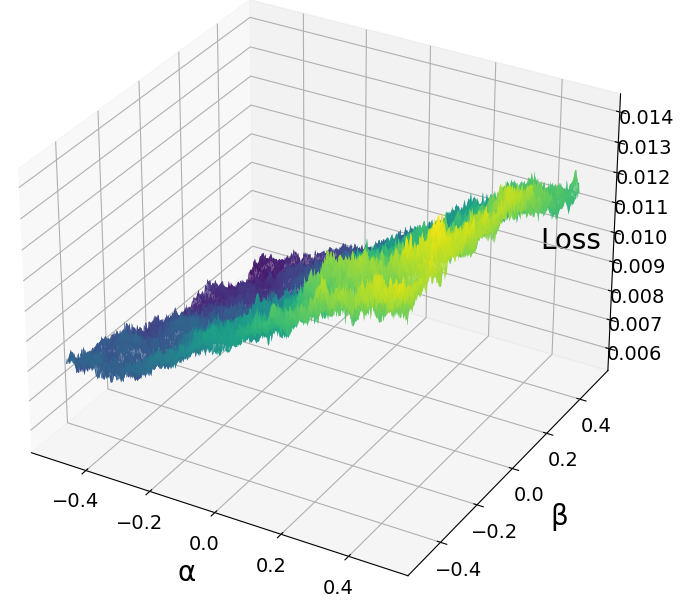}%
            \label{subfig:random-loss}%
        }%
        \caption{Comparison of 3D loss landscape visualization methods: (a) uses the top-2 eigenvectors of model parameters, while (b) uses two random orthogonal directions.}%
        \label{fig:hessian-vs-random}%
\end{figure}

However, when considering models with thousands of parameters, picking one random direction may lead to a gross approximation of the loss landscape which is not practically representative (e.g., Figure \ref{subfig:random-loss}). For this reason, we propose a novel approach that selects $\sigma$ and $\eta$ as the directions of the top-2 eigenvectors of the converged model. In this way, we can explore the two directions where the loss landscape faces the maximum curvature, leading to a more informative topology approximation of the surroundings of the model parameters $\theta$ (e.g., Figure \ref{subfig:hessian-loss}). A detailed comparison of the two approaches is discussed in Appendix~\ref{sec:abl-vis}.

\subsection{CKA Similarity}
In the context of loss landscape analysis,
the Centered Kernel Alignment (CKA) similarity \cite{cka2,  cka} is used to determine whether multiple instances of the same model, trained separately with randomly initialized parameters, tend to converge towards similar minima. This is assessed by examining the CKA similarity of their outputs, i.e., given NN \( f_{\theta} \) and $m \in \mathbb{N}$ data samples randomly picked from the test set, we can define their concatenation as \( F_{\theta} = [f_{\theta}(x_1) \cdots f_{\theta}(x_m)]^T \in \mathbb{R}^{m \times d_{\text{out}}} \). 

The CKA similarity between two NNs with different parameter configurations \( \theta \) and \( \theta' \) can hence be measured by
\begin{equation}\label{eqn:cka}
    \text{CKA}(\theta, \theta') = \frac{\text{Cov}(F_{\theta}, F_{\theta}')}{\sqrt[]{\text{Cov}(F_{\theta}, F_{\theta})\text{Cov}(F_{\theta}', F_{\theta}')}}.
\end{equation}
    
In the above equation, given two matrices $ X, Y \in \mathbb{R}^{m \times d}$, it holds
\begin{equation}\label{eqn:cka-2}
\text{Cov}(X, Y) = (m - 1)^{-2}tr(X X^T H_m Y Y^T H_m),
\end{equation}
where $H_m = I_m - \frac{1}{m}\textbf{11}^T$ is the centering matrix.\footnote{We use $I_m$ to denote the identity matrix of size $m$ and $\textbf{1}$ represents the \textit{indicator function}.} Further details are discussed in the Appendix~\ref{sec:abl-cka}.

A high CKA value indicates that the models are likely converging closely to each other, whereas a low value suggests that different parameter initializations can lead to convergence in different regions, as also empirically demonstrated by \cite{loss_landscape}. This information helps describe the morphology of the loss landscape: in a globally smooth and flat loss landscape, models initialized with different parameters are expected to converge to similar, nearby minima. In contrast, low similarity may indicate a rugged loss landscape, where models risk getting trapped in suboptimal local minima.


\subsection{Hessian Metrics}
The Hessian is a square matrix that characterizes the curvature of the loss function at a specific point. The eigenvalues of the Hessian provide scalar values that offer insights on the curvature type at that point. Positive eigenvalues suggest local convexity of the loss, indicating a single minimum or maximum. In contrast, negative eigenvalues suggest local concavity, which implies the presence of a saddle point, often resulting from unfavorable training conditions. Zero eigenvalues signify a flat loss at that point, indicating the absence of both a minimum and a maximum. Intuitively, it is desirable to have both the top eigenvalue and the sum of the traces as close to zero as possible, because it suggests that the model has converged to a smooth, flat minimum.

Computing the Hessian matrix can be challenging from a computational perspective. In fact, it involves evaluating second-order partial derivatives of the loss function with respect to each pair of the model parameters. However, our analysis focuses on models deployed on edge devices, which are characterized by a limited number of parameters compared to foundational models. Furthermore, we leverage PyHessian \cite{pyhessian} to calculate Hessian metrics, an open-source framework that approximates Hessian values by applying power methods.

\subsection{Mode Connectivity}\label{sec:mc}
Mode connectivity is a global metric that provides insights about how well two minima are connected \cite{mode_connectivity, mode_connectivity2}. It is capable of revealing the presence of barriers of loss between two points in which the two models converged. 
A simple way to compute mode connectivity is to set up a linear interpolation between two given models and sample a certain number of parameter configurations along it. Then, we can compute the loss of the models resulting from the sampled parameter configurations and see if there are barriers. However, this is a gross approximation because we do not know a priori the shape of the minima: for instance, it may happen that the two models are well connected via a curved line, whereas the linear interpolation cannot catch this information. A more convoluted method is hence required. We adopted the one proposed by \cite{mode_connectivity}, which is based on a parameterized Bezier curve with $k+1$ bends. Consider $k+1$ models with parameters $\phi = \{\theta_0, \ldots, \theta_k \}$, where $\theta_0 = \theta'$ and $\theta_k = \theta''$ are the parameters of the two models to be compared, while $\theta_j$ for $1 \leq j < k$ are the parameters of other models to be trained. The Bezier curve is defined as:
\begin{equation}\label{eqn:bezier}
    \gamma_{\phi}(t) = \sum_{j=0}^{k} \binom{k}{j} (1 - t)^{k - j} \cdot t^j  \cdot \theta_j,
\end{equation}
where $t \in [0,1]$ is a scalar to move along the curve.
The training of the models related to $\theta_j$ for $1 \leq j < k$ is performed as follows: first, the parameters are initialized using a linear interpolation between $\theta_0$ and $\theta_k$; then each of such models is trained to reach convergence.

At this point, we can now sample $m >2$ different parameter configurations by picking $m$ values of $t$, denoted by $\mathcal{T} = \{ t_i \}_{i=0}^{m-1}$, 
including the two boundaries $t_1=0$ and $t_{m-1}=1$. Intermediate values are computed by $t_i = i/(m-1),$ for $i = 1, \ldots, m-2$.

 We define the distance between the average loss of the two models being compared and the loss of a sampled one by
 \begin{equation}\label{eqn:mc-dist}
 d(t, \theta', \theta'') = \frac{1}{2}(\mathcal{L}(\theta') + \mathcal{L}(\theta'')) - \mathcal{L}(\gamma_{\phi}(t)).
 \end{equation}
Finally, we can now define mode connectivity as the maximum deviation from the average loss of the two boundaries in the selected sampling points $t_i$, i.e., 
$mc(\theta', \theta'') = d(t^*)$
where $t^* = \argmax_{t \in \mathcal{T}} \{ |d(t, \theta', \theta'')| \}$.

%
%
%
%
This metric can be interpreted as follows:
\begin{equation*}
    \begin{cases}
    mc(\theta', \theta'') > 0, & \text{ There are better minima.}\footnotemark[2] \\
    mc(\theta', \theta'') < 0, & \text{ There are barriers.} \\
    mc(\theta', \theta'') \approx 0, & \text{ Loss landscape is well connected.}
    \end{cases}
\end{equation*}
\footnotetext[2]{The training failed to locate a reasonable optimum, i.e., $\mathcal{L}(\theta')$ and $\mathcal{L}(\theta'')$ are large.}
In this work, we computed mode connectivity with 3 bends ($k=2$), training the models used to shape the Bezier curve for 30 epochs.
A graphical example of mode connectivity is available in Appendix~\ref{sec:mc_ex}, while an ablation study about how we set the right number of bends and training epochs is presented in Appendix~\ref{sec:mc-abl}.

\section{Experimental Setup}
\label{sec:setup}
This section provides a detailed overview of the models employed in this study as representative benchmarks, and the noise mitigation techniques we tested.

\subsection{Benchmark Models}
Both models employed in our analysis are used for scientific sensing and are designed to be deployed on resource-constrained devices such as Field Programmable Gate Arrays (FPGAs) and Application Specific Integrated Circuits (ASICs), where the number of model parameters and the architectural design play a pivotal role in achieving efficiency.

\subsubsection{ECON-T model}
The ECON-T model introduced by \cite{econ} is an autoencoder for lossy data compression created for the Large Hadron Collider (LHC) and its high luminosity upgrade (HL-LHC) at CERN. 
Figure~\ref{subfig:econ-fig} shows an example compression flow employed by ECON-T. We focus our analysis on the encoder composed of 2288 parameters, and deployed on the detector in a high-radiation environment. The size and complexity of the model are constrained by area, on-chip memory, and power ($\leq 100$ mW). The performance of the autoencoder is measured via Earth Mover’s Distance (EMD)~\cite{emdimage}, which is not differentiable, 
so it is not used during training. A physics-inspired loss similar to Mean Square Error (MSE) called ``telescoping MSE'' is used during training, and EMD is used for measuring performance. 
Differentiable EMD loss~\cite{telescope_loss} has been studied but is not benchmarked here.  

\subsubsection{Fusion model}
Active feedback control in thermonuclear fusion devices based on magnetic confinement is required to mitigate plasma instabilities and enable robust operation, preventing damage to the reactor. \cite{Wei:2023mma} combined efficient processing of FPGAs with high-speed imaging camera diagnostic and convolutional neural networks (CNNs) for magnetohydrodynamic (MHD) mode control on a tokamak device. The workflow of this system is illustrated in Figure \ref{subfig:fision-fig}. The model inputs a camera image and predicts the $n=1$ MHD mode amplitude and phase, where $n$ is the \textit{number density} used to describe the degree of concentration of countable objects 
in physical space. As CNN are not the SoTA for phase predictions due to periodicity of the task, between $-\pi$ and $\pi$, this work focuses on amplitude only.

\subsection{Scientific Corruptions}
Neural network quantization may not be the only noise facing the model at the stage of deployment. Indeed, models used in scientific experiments may operate in harsh environments, such as space or particle accelerators, where they typically experience significant performance degradation due to noise in the input data and weight perturbation, such as Single Event Upsets (SEU). To simulate the perturbations in the input data we adopt the injection of two different types of noise:
\begin{itemize}[]
    \item \textbf{Gaussian noise}: It appears as random variations in pixel intensity that follow a Gaussian distribution. It typically arises from electronic noise in imaging sensors or during transmission.
    \item \textbf{Salt and Pepper noise}: It commonly arises from defects in imaging sensors, transmission errors, or faulty pixels in digital cameras. Unlike Gaussian and random noise, salt-and-pepper noise introduces localized disruptions in image content, which can severely degrade image quality. The target pixels are randomly selected without following any particular distribution and their value is either maximized or minimized.
\end{itemize}

We evaluated the performance of the models under varying levels of noisy perturbations to better understand their sensitivity to distorted input data. 
It is important to note that, in a practical setting, the nature of the perturbation is typically unknown a priori. 
As we will demonstrate later, in section \ref{sec:results}, in such cases, mitigation techniques that are agnostic to the type of noise are generally more effective.

In addition to input corruptions, we also study weight corruptions from SEUs. On the software side, we adopted the FKeras~\cite{fkeras} methodology, which ranks bits approximately from most to least sensitive to flipping. This allows us to simulate worst-case scenarios by flipping the top-$k$ most sensitive bits and then evaluating the model's performance under these conditions. The sensitive bit ranking is done by first sorting the weights by a sensitive score computed as
$
    H' = \sum_{i = 1}^k \lambda_i (v_i \cdot \theta) v_i ~~ \in \mathbb{R}^{n},
$
where $k$ indicates the number of top eigenvectors of the model, $\lambda_i$ is the i-th eigenvalue, $v_i$ is the i-th eigenvector of the model, and $\theta$ is the vector of model parameters ($n$ is the number of parameters). Once the parameters are sorted by sensitivity, we then sort the bits of each parameter from the most significant bit (MSB) to the least significant one (LSB). 

In Appendix~\ref{sec:bench_abl}, we demonstrate the effectiveness of FKeras with respect to random bit flipping. 

\textit{Note}: Adversarial attacks were not considered, as models used in scientific experiments
are deployed in controlled environments where they 
are not exposed to this threat.

\begin{figure*}[ht!]
        \subfloat[]{%
            \includegraphics[width=0.25\linewidth]{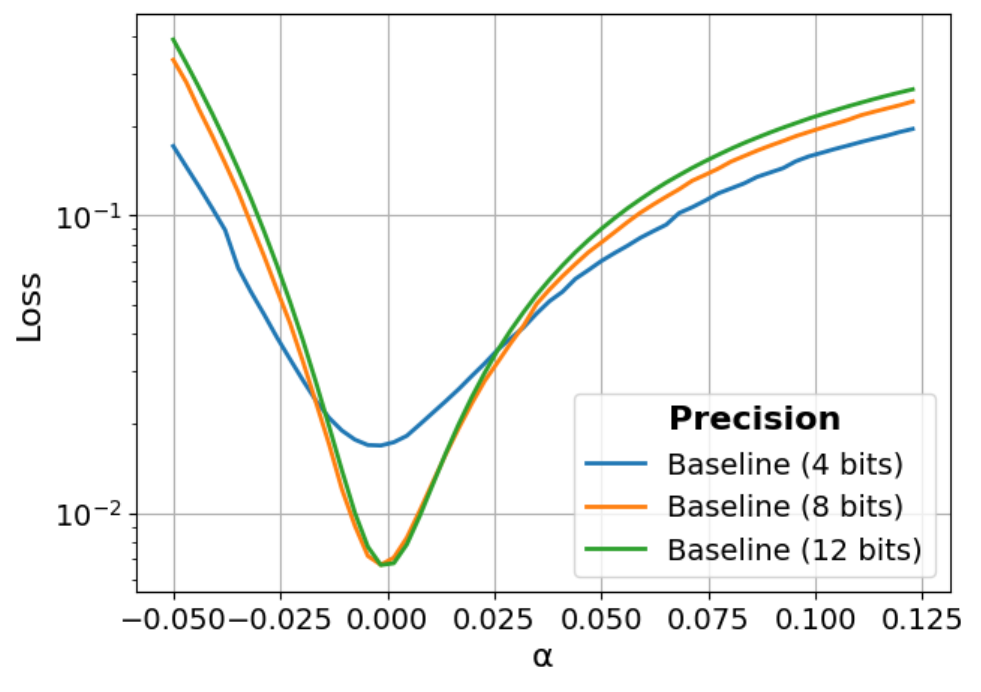}%
            \label{subfig:econ-loss-precision}%
        }
        \subfloat[]{%
            \includegraphics[width=0.25\linewidth]{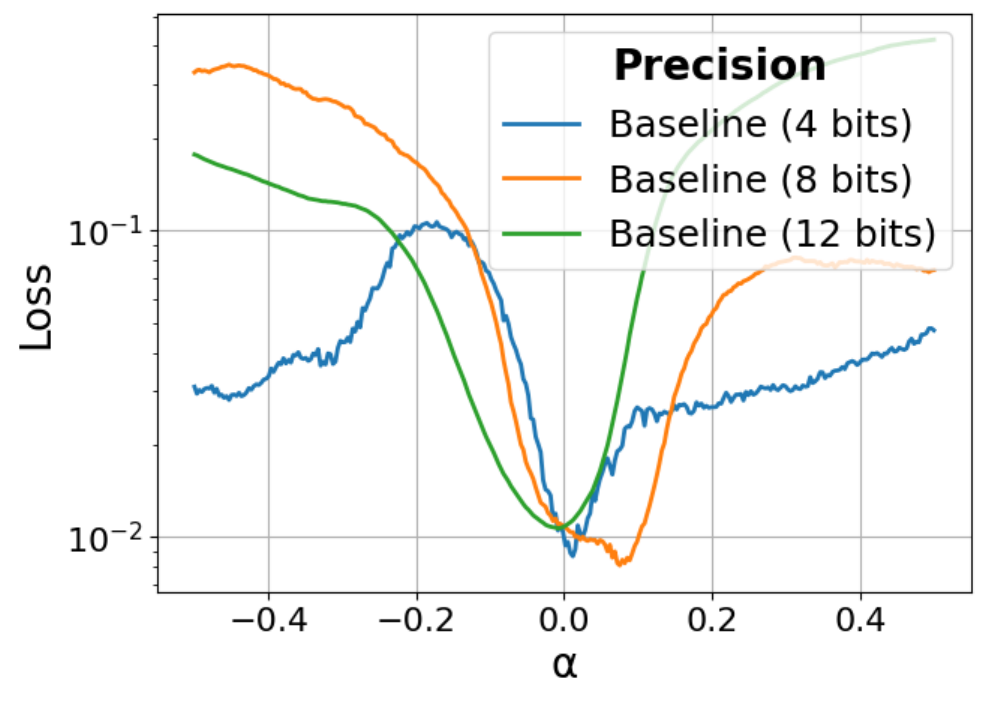}%
            \label{subfig:fusion-loss-precision}%
        }
        \subfloat[]{%
            \includegraphics[width=0.25\linewidth]{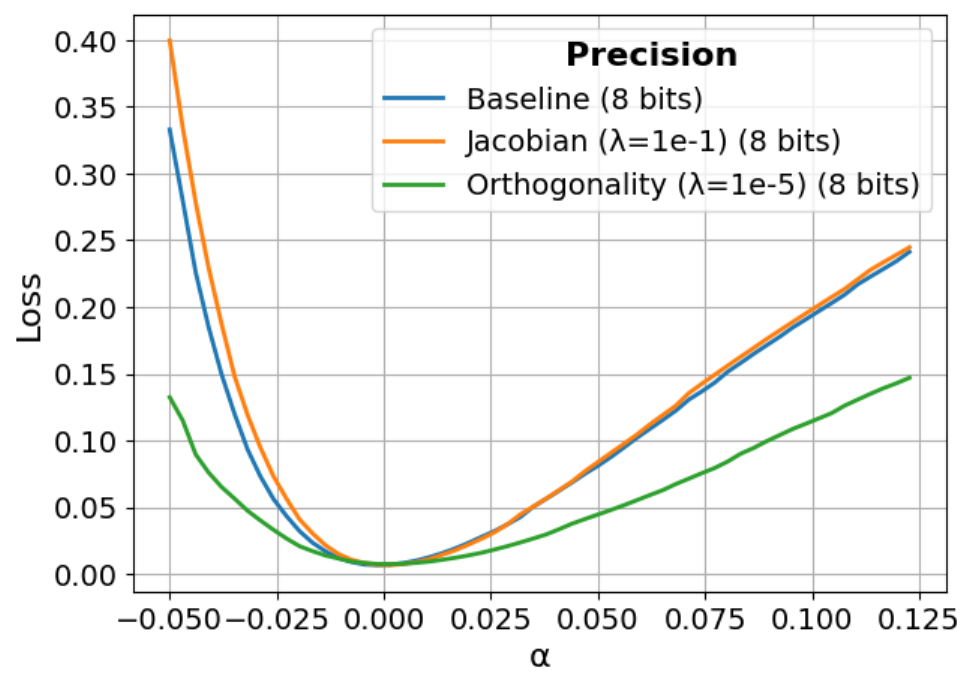}%
            \label{subfig:econ-loss-reg}%
        }
        \subfloat[]{%
            \includegraphics[width=0.25\linewidth]{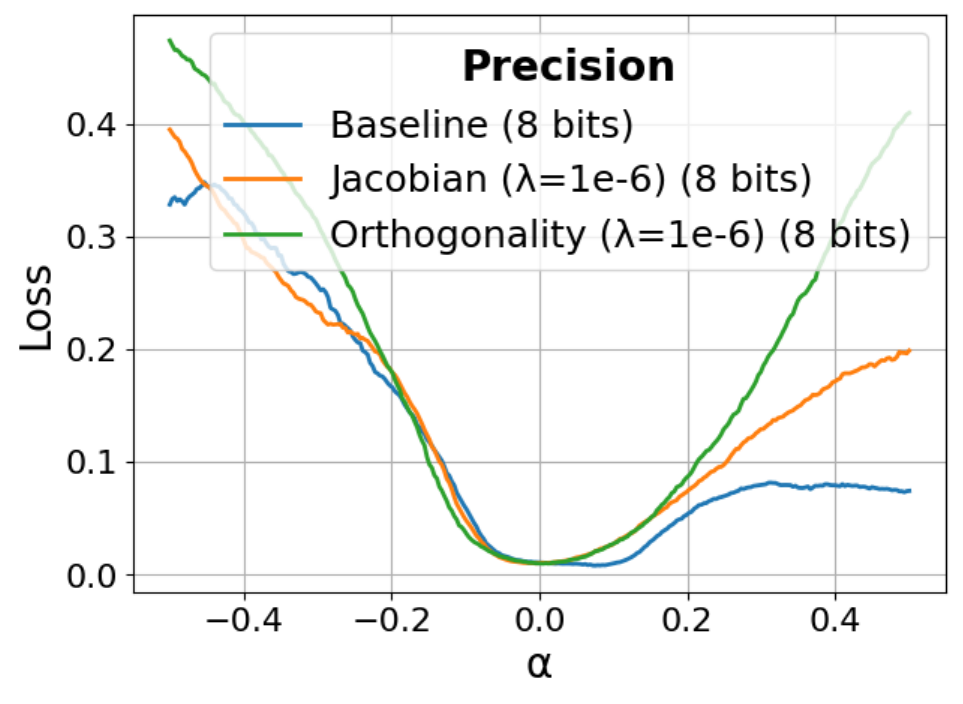}%
            \label{subfig:fusion-loss-reg}%
        }
        
        \caption{Comparison of loss curves computed by perturbing the models along the top eigenvector of the Hessian matrix (i.e., varying parameter $\alpha$ in Eq.~\eqref{eqn:loss-land}, reported on the x-axis, while keeping $\beta=0$). (a) and (b) compare the loss line of models trained with different precision respectively for ECON-T and Fusion models, while (c) and (d) compare models fine-tuned with different regularization techniques respectively for ECON-T and Fusion models.}
        \label{fig:visualization}
\end{figure*}

\subsection{Noise mitigation methods}
To demonstrate the correlation between loss landscape analysis and robustness, we evaluated different versions of the target models, trained with different regularization methods. These methods aim to increase the robustness of the models against input and parameter perturbations by modifying the model's loss function and thus also its loss landscape. However, we will see in section \ref{sec:results}, that they are not always effective, and how the morphology of the loss landscape can help us to understand which one is more effective.

In this work, we compare two noise mitigation techniques: Jacobian regularization \cite{jacobian2, jacobian} and orthogonal regularization \cite{parseval, orthCNN, orthQNN}.

Jacobian regularization aims to limit the impact of input perturbations by adding a penalty to the loss function, which is proportional to the Frobenius norm (denoted by $\|\cdot\|_F$) of the model’s Jacobian matrix $J(x)$, i.e.,
\begin{equation}\label{eqn:reg-jac}
    \delta \| J(x)\|_F^2,
\end{equation}
where $\delta$ is a scalar used to weigh the impact of the regularization on the training loss.

This method controls the magnitude of the components of the Jacobian matrix, which are partially correlated with the contribution of noise to the model output. The aim of this method is to increase the margin between the input space and the decision boundaries of the target class. In our experiments, we use an efficient approximation of the Frobenius norm of the Jacobian matrix proposed by \cite{jacobian}, which allows us to implement this method with negligible overhead.

The relationship between orthogonality and quantization has previously been studied by \cite{orthQNN}. Although they demonstrated the beneficial effects of enforcing orthogonality among neural network weights during QAT, our goal is to provide a more detailed analysis of the effect of orthogonality on the loss landscape and reveal a possible correlation of it with a model's robustness. Various approaches to promote weight orthogonality have been proposed in the literature \cite{orth2018, orth2018b, orth2018c, orthCNN}. In this work, we use a soft orthogonal regularization based on the Frobenius norm, formulated as:
\begin{equation}\label{eqn:reg-ort}
    \delta \| W^T W - I \|_F.
\end{equation}
More complex regularization formulations could be applied, but given the nature of the model and the objectives of this study, this technique provides sufficiently effective results. 

Moreover, in this work we did not take into account defensive approaches such as adversarial training \cite{adv_training} or noise injection during training for mainly two reasons: first, we have no guarantees that these methods will enhance reliability against noise of different nature; and second, the overhead introduced during training it is not negligible, especially when combined with QAT. 

In Appendix~\ref{sec:reg_abl}, we study how different values of $\delta$ impact the performances of the model.

\section{Experimental Results}
\label{sec:results}
In this section, we first present results to evaluate the metrics introduced above and then assess the potential correlation between these metrics and the robustness of the tested models (ECON-T and Fusion). We studied robustness in the presence of network quantization. The models were quantized using integer uniform quantization implemented in Brevitas~\cite{brevitas} library, and all experiments were conducted on an NVIDIA A100 GPU using QAT. We evaluated \emph{three versions} of each model: \textbf{(i)} a baseline version fine-tuned without regularization, \textbf{(ii)} one incorporating \textit{Jacobian regularization} (with $\delta = 0.1$ for the ECON-T Model and $\delta = 10^{-6}$ for the Fusion model), and \textbf{(iii)} another one employing \textit{orthogonal regularization} (with $\delta = 10^{-5}$ for the ECON-T Model and $\delta = 10^{-6}$ for the Fusion model). This approach assesses how a quantization scheme and noise mitigation influence the performance and reliability of the models. Each model version was then trained three times under different precisions, i.e., for bit widths ranging from 3 to 12. Unless otherwise stated, the results presented in this section represent the average of these three model versions. 

Codes are available at \url{https://github.com/balditommaso/PyLandscape}.

\begin{figure}[ht!]
        \centering
        \includegraphics[width=.6\linewidth]{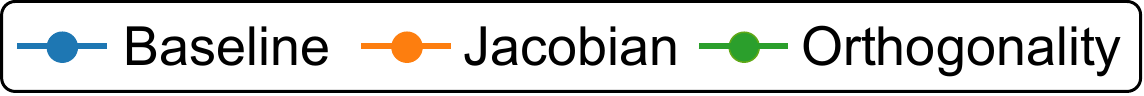}
        \subfloat[]{%
            \includegraphics[width=0.49\linewidth]{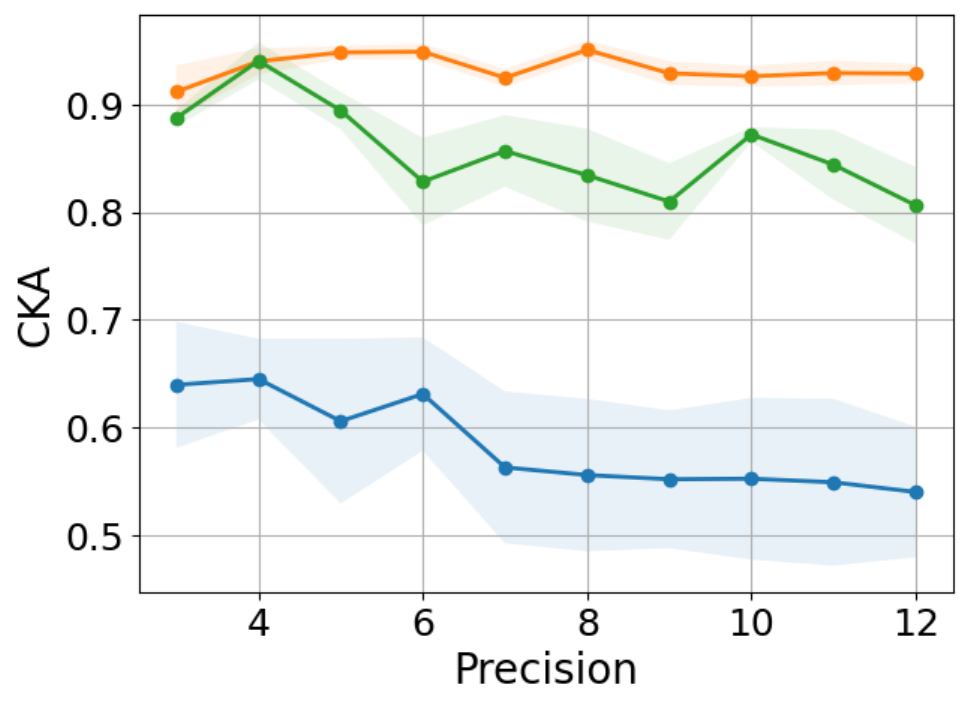}%
            \label{subfig:econ-cka}%
        }%
        \subfloat[]{%
            \includegraphics[width=0.5\linewidth]{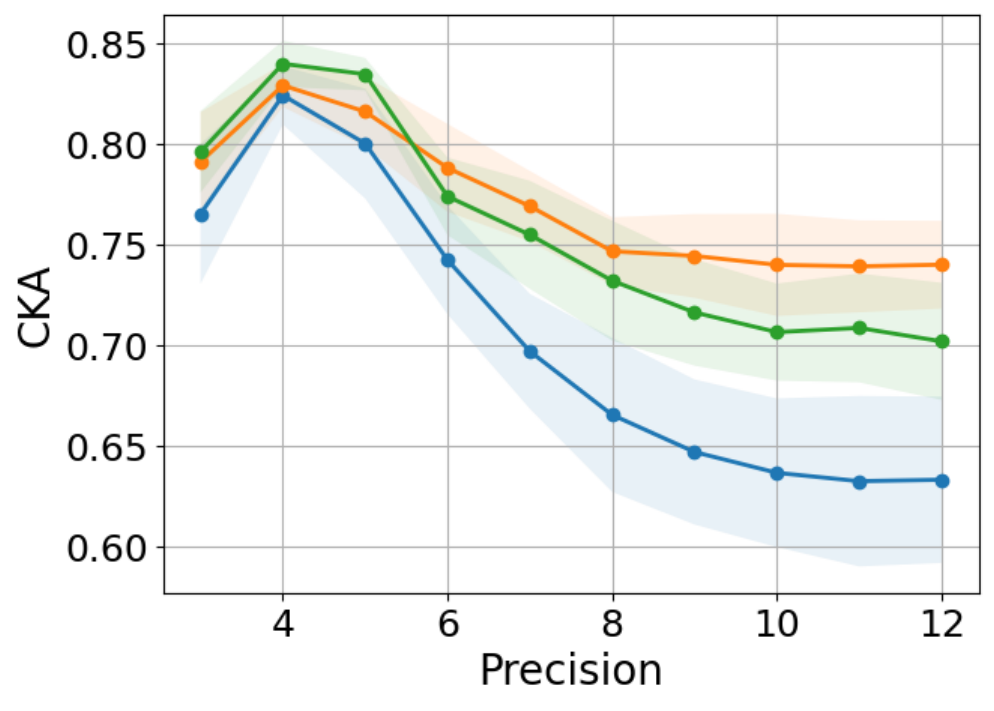}%
            \label{subfig:fusion-cka}%
        }\\%
        \subfloat[]{%
            \includegraphics[width=0.5\linewidth]{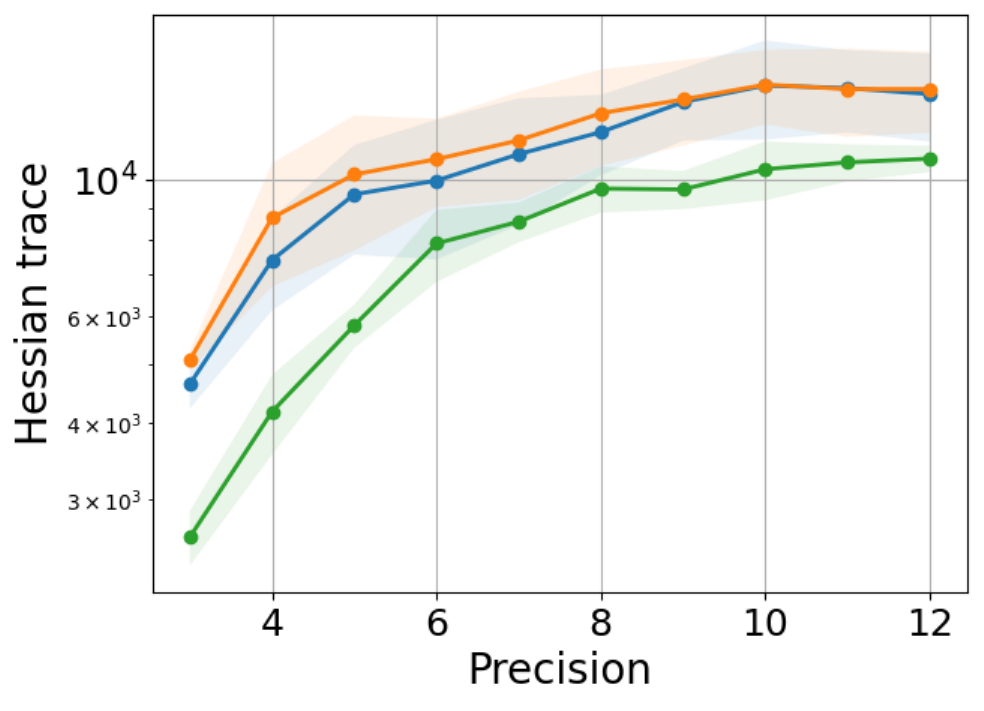}%
            \label{subfig:econ-trace}%
        }%
        \subfloat[]{%
            \includegraphics[width=0.5\linewidth]{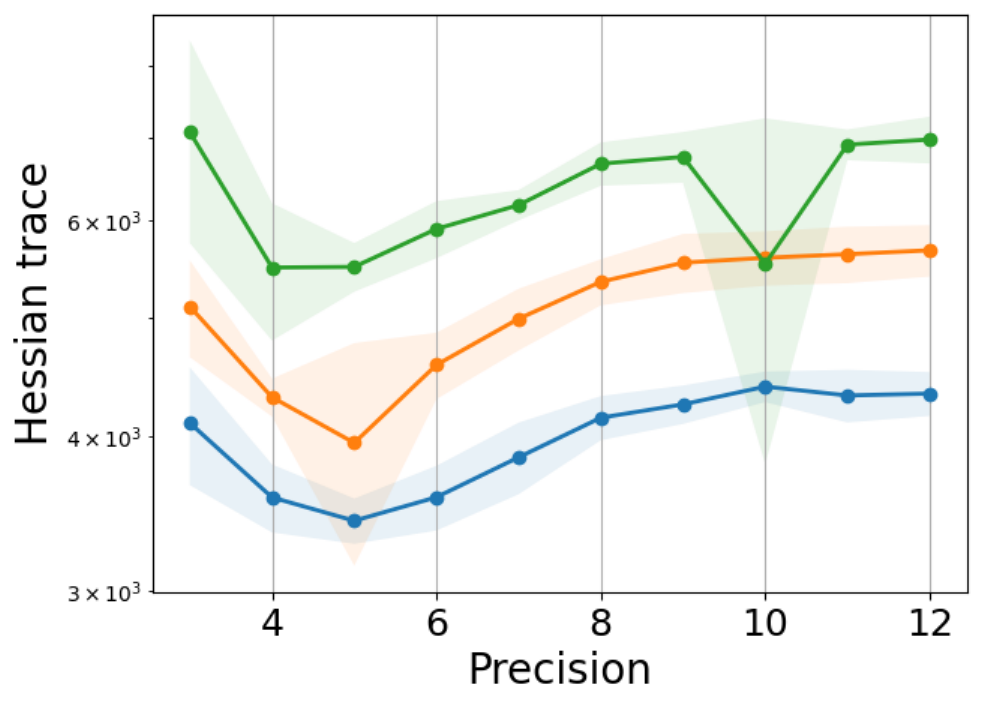}%
            \label{subfig:fusion-trace}%
        }\\%
        \subfloat[]{%
            \includegraphics[width=0.48\linewidth]{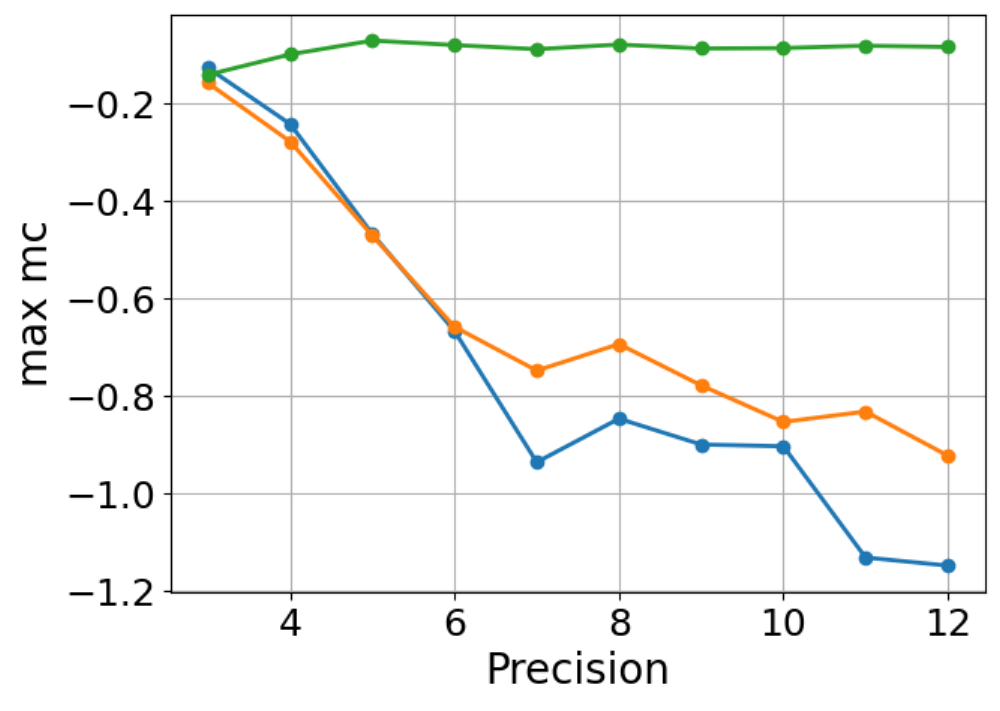}%
            \label{subfig:econ-mc}%
        }
        \subfloat[]{%
            \includegraphics[width=0.5\linewidth]{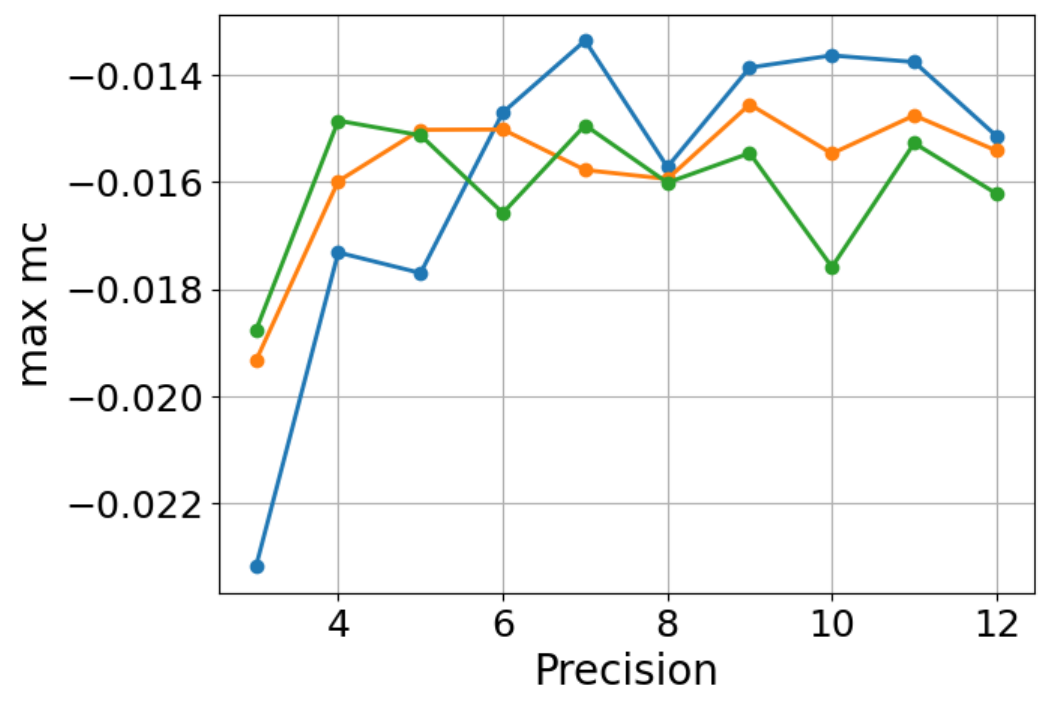}%
            \label{subfig:fusion-mc}%
        }
        \caption{Analysis of loss landscape metrics for ECON-T (left column) models and Fusion models (right column) fine-tuned with different regularization strategies across varying precision levels. Subplots show: (a) and (b) \textbf{CKA similarity}, which evaluates representational alignment among models; (c) and (d) \textbf{Hessian trace}, capturing the overall curvature of the loss landscape where the model is converged; and (e) and (f) \textbf{mode connectivity}, indicating the presence of barriers among different minima. Regularization methods include Baseline (no regularization), Jacobian regularization, and orthogonal regularization. 
        }%
        \label{fig:benchmarks}%
\end{figure}

\subsection{Loss Landscape analysis}
\textbf{Visualizing the Loss Landscape.}
Visualization techniques provide an approximate representation of the shape of the loss landscape. In this work, we utilized 2D plots.
The intrinsic regularization effect of quantization is illustrated in Figure \ref{subfig:econ-loss-precision}. The 4-bit version of the baseline model exhibits a higher minimum compared to the 8-bit and 12-bit configurations, due to the performance degradation typically associated with low-bit quantization. However, examining the entire loss curve reveals that the convex portion of the loss landscape is wider and flatter compared to models trained with higher precision. In contrast, Figure \ref{subfig:fusion-loss-precision} demonstrates that low-precision quantization achieves a minimum comparable to the one of higher-precision configurations, but at the cost of sharper and more jagged minima. This results in a harsher loss landscape, which can complicate model training. As also discussed later, this has a significant impact on model robustness. For the ECON-T model, orthogonal regularization proves to be more effective than Jacobian regularization, resulting in a smoother and wider convex loss landscape (Fig. \ref{subfig:econ-loss-reg}). Conversely, for the Fusion model, the impact of regularization on the loss landscape is limited (Fig. \ref{subfig:fusion-loss-reg}).

\textbf{CKA similarity.} We are interested in understanding if different instances of the same model converge in a close area of the loss landscape by looking at their CKA similarity. Figures \ref{subfig:econ-cka} and \ref{subfig:fusion-cka} report the average CKA similarity between all pairs of parameter configurations (resulting from the three trainings) of the three model versions as a function of the quantization precision in bits. The figures show that the baseline versions have limited CKA similarity, which tends to decrease as the precision increases, probably due to the implicit regularization effect resulting from low-bit quantization. This is not the case for the other two model variants, where, especially for Jacobian regularization, they converge to CKA-similar models. In these cases, the results suggest the presence of a smoother loss landscape, where the models do not get trapped in suboptimal minima during training. The drop in CKA similarity as the precision increases is more significant for the Fusion model (Fig.~\ref{subfig:fusion-cka}) than the ECON-T model (Fig.~\ref{subfig:econ-cka}): this was expected because the Fusion model has more parameters, leading to a more complex loss landscape).

\textbf{Hessian trace.} The curvature of the loss landscape where the model lands at the end of QAT can be analyzed with the Hessian trace. The average Hessian trace values are reported in Figures~\ref{subfig:econ-trace} and~\ref{subfig:fusion-trace}. From the figures we can see that all models tend to follow the same pattern in which the slope of the loss landscape increases with precision $\geq$ 5 bits. 
Furthermore, while the ECON-T model (inset (c)) benefits from orthogonal regularization, which allows converging to flatter minima, regularization is instead letting the Fusion model (inset (d)) converge to steeper minima. 
These behaviors are further investigated later in this section.

\textbf{Mode connectivity.} The presence of barriers in the loss landscape is undesirable as they hinder the optimization process, making it difficult for algorithms to efficiently converge to a global or near-global minimum. Weight perturbations caused by bit errors can shift the model's position in the loss landscape, with the magnitude of the shift determined by the difference between the original and perturbed parameters and the influenced direction. Intuitively, if the region surrounding the model is free of barriers, the impact of perturbations on performance is likely to be lower. We studied the maximum mode connectivity (Max mc) obtained as follows. Given the three models for each version, we sampled $m=60$ points on the corresponding Bezier curve (Eq.~\eqref{eqn:bezier}) and denote by $T$ the set of model parameters corresponding to those 60 points. Max mc is hence given by the maximum mode connectivity of all pairs in $T$, i.e., $\max_{(\Theta', \Theta'') \in T \times T} \{mc(\Theta', \Theta'') \}$. Figures \ref{subfig:econ-mc} and \ref{subfig:fusion-mc} illustrate how the presence of barriers is influenced by precision. Note that not all regularization methods effectively mitigate these barriers. The two models exhibit different behaviors: for ECON-T (Fig. \ref{subfig:econ-mc}), Jacobian regularization provides only slight improvements in connectivity for precisions ranging from 6 to 12 bits, whereas models fine-tuned with orthogonal regularization are well connected, demonstrating the absence of significant barriers. In contrast, for Fusion (Fig. \ref{subfig:fusion-mc}), the presence of barriers between minima decreases as precision increases, with regularization mainly offering benefits in low-precision configurations. 

\subsection{Performance under perturbations}
\begin{figure*}[ht!]
        \centering
        \includegraphics[width=0.6\linewidth]{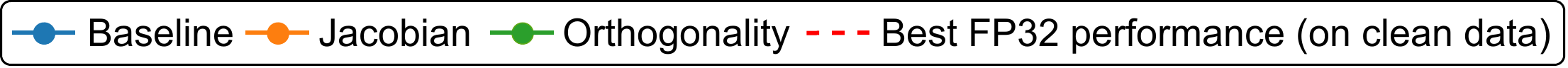}
        \subfloat[]{%
            \includegraphics[width=0.25\linewidth]{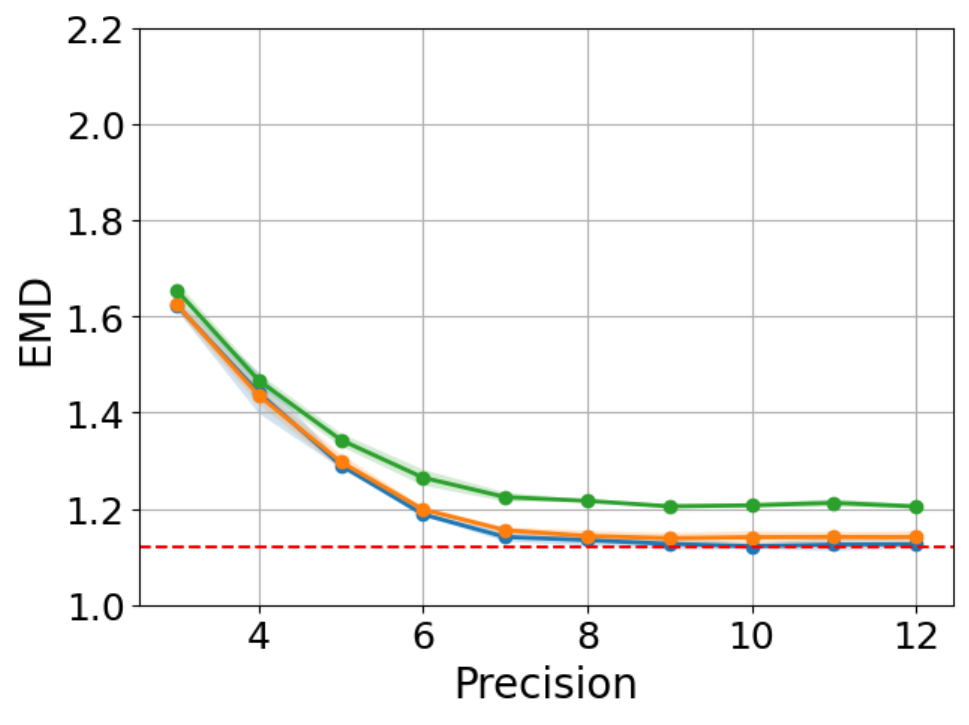}%
            \label{subfig:econ-clean}%
        }%
        \subfloat[]{%
            \includegraphics[width=0.25\linewidth]{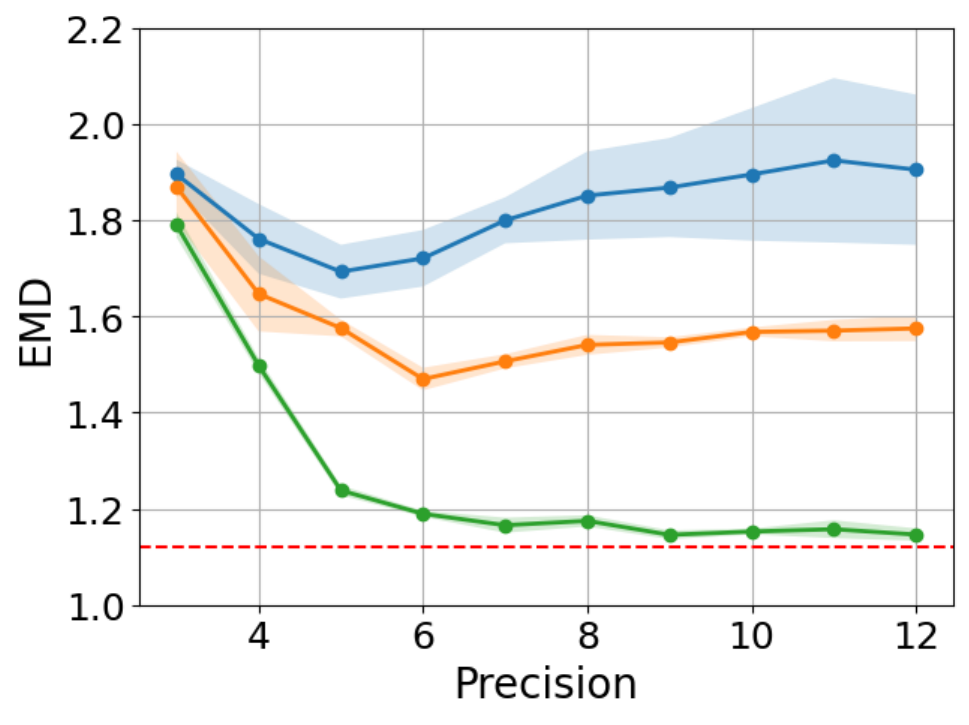}%
            \label{subfig:econ-gaussian}%
        }%
        \subfloat[]{%
            \includegraphics[width=0.25\linewidth]{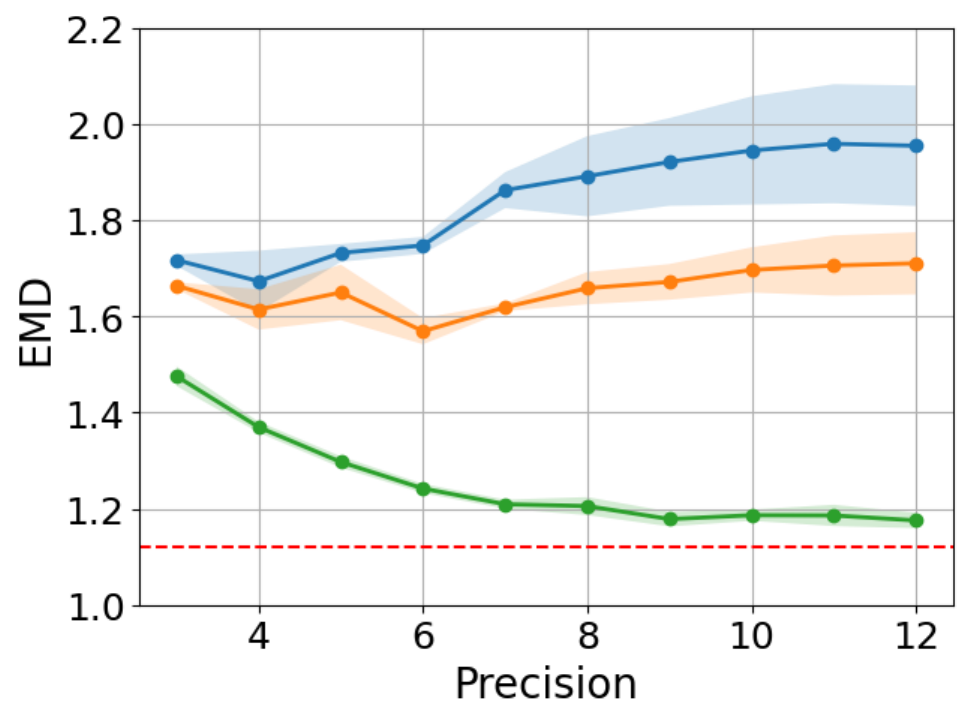}%
            \label{subfig:econ-s_p}%
        }%
        \subfloat[]{%
            \includegraphics[width=.25\linewidth]{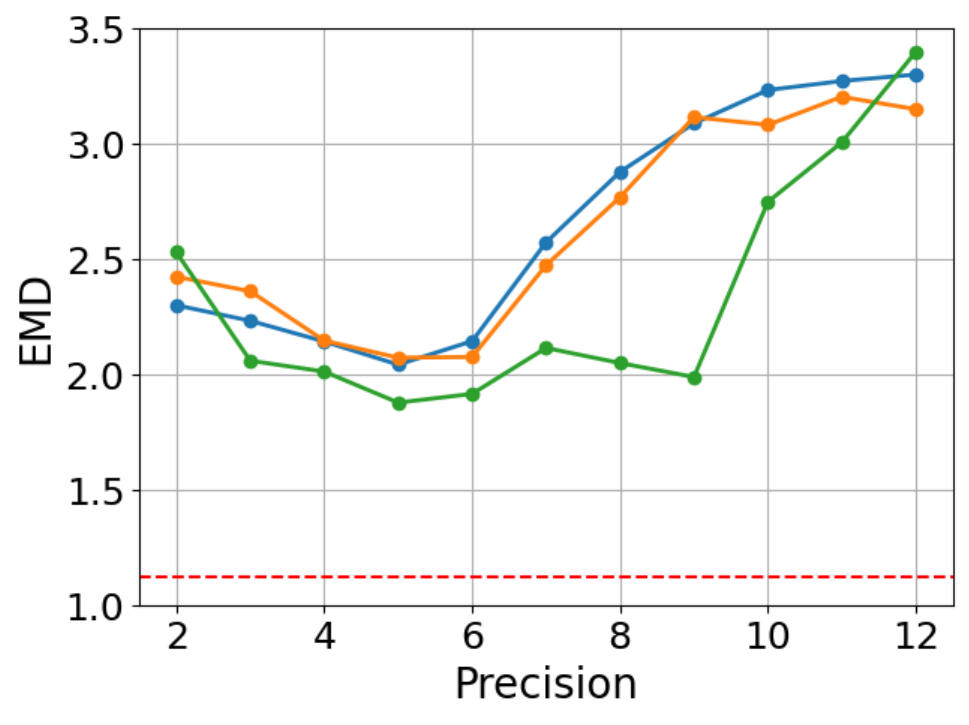}%
            \label{subfig:econ-fkeras}%
        }\\%
        \subfloat[]{%
            \includegraphics[width=0.25\linewidth]{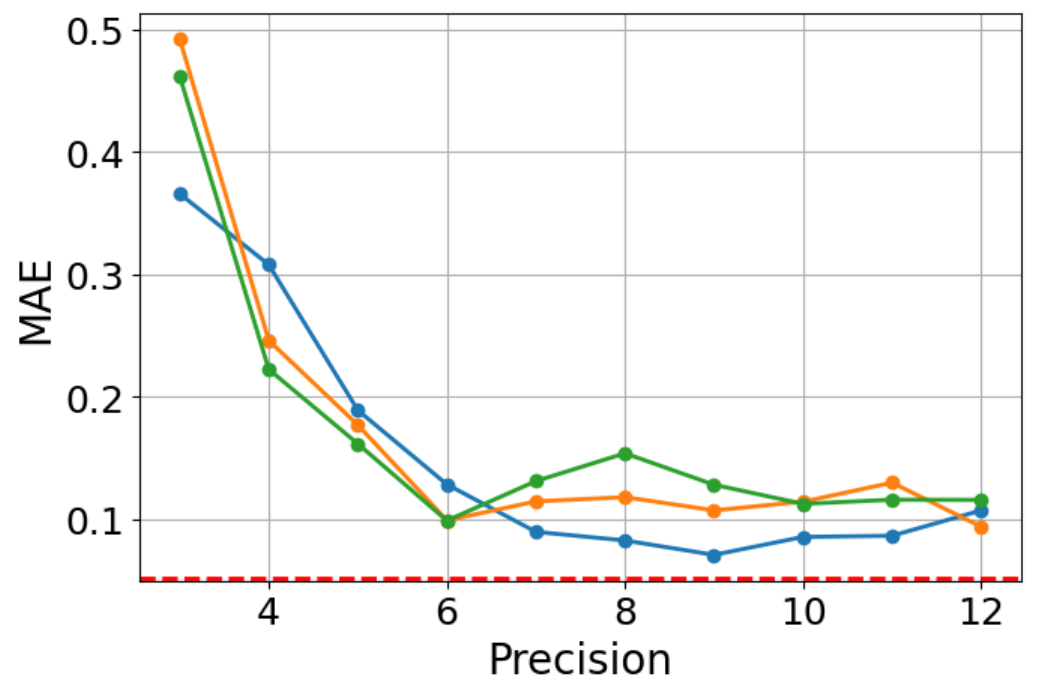}%
            \label{subfig:fusion-clean}%
        }%
        \subfloat[]{%
            \includegraphics[width=0.25\linewidth]{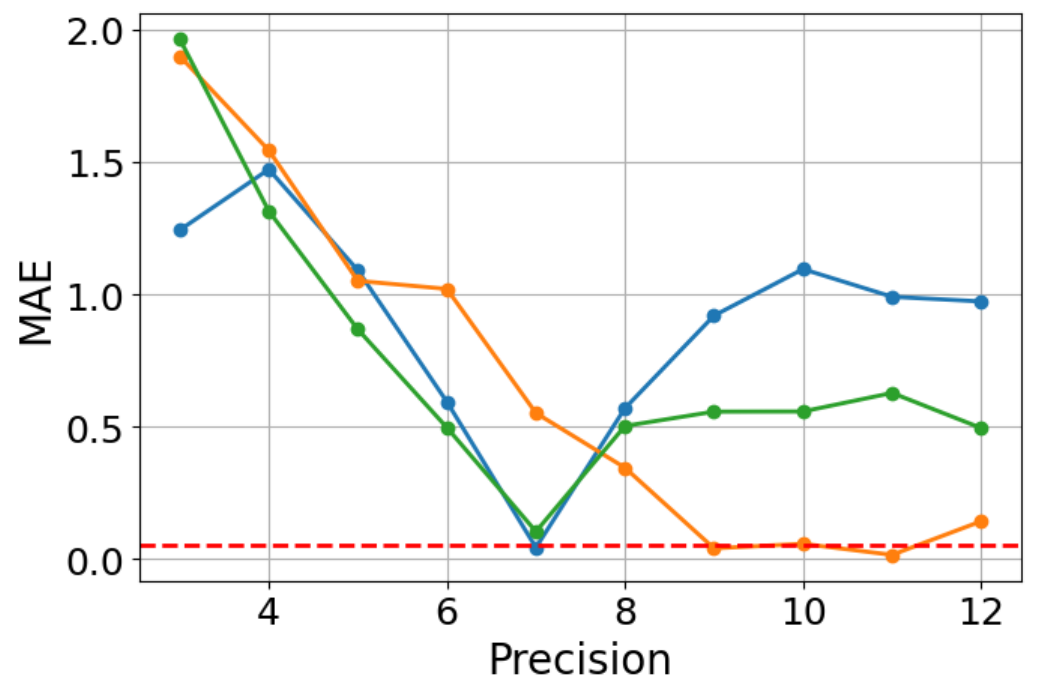}%
            \label{subfig:fusion-gaussian}%
        }%
        \subfloat[]{%
            \includegraphics[width=0.25\linewidth]{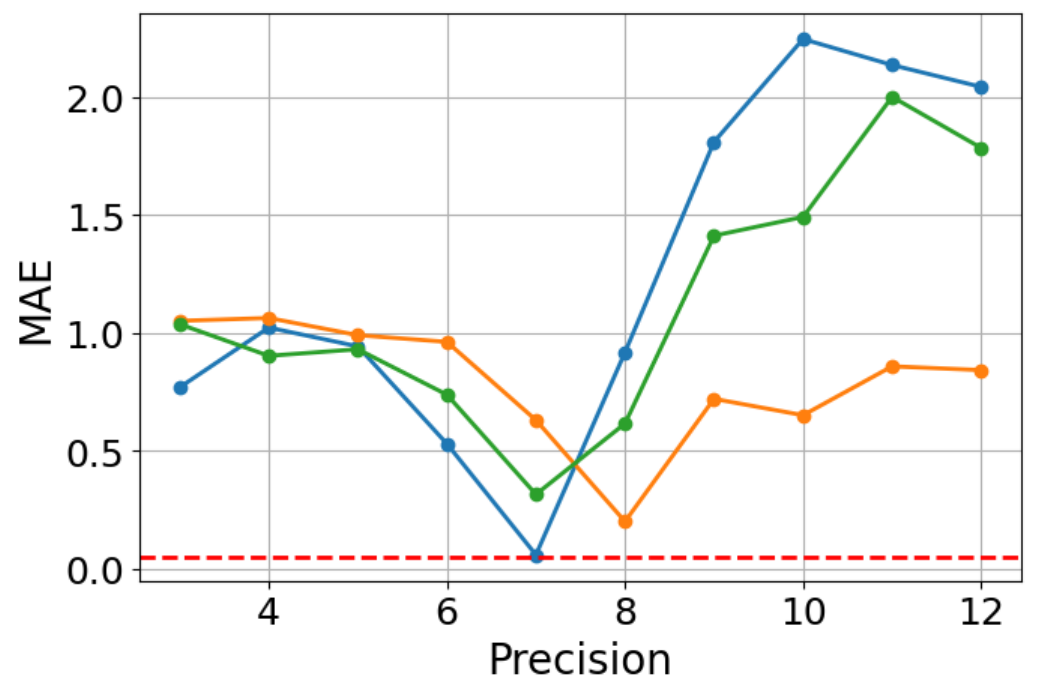}%
            \label{subfig:fusion-s_p}%
        }%
        \subfloat[]{%
            \includegraphics[width=.25\linewidth]{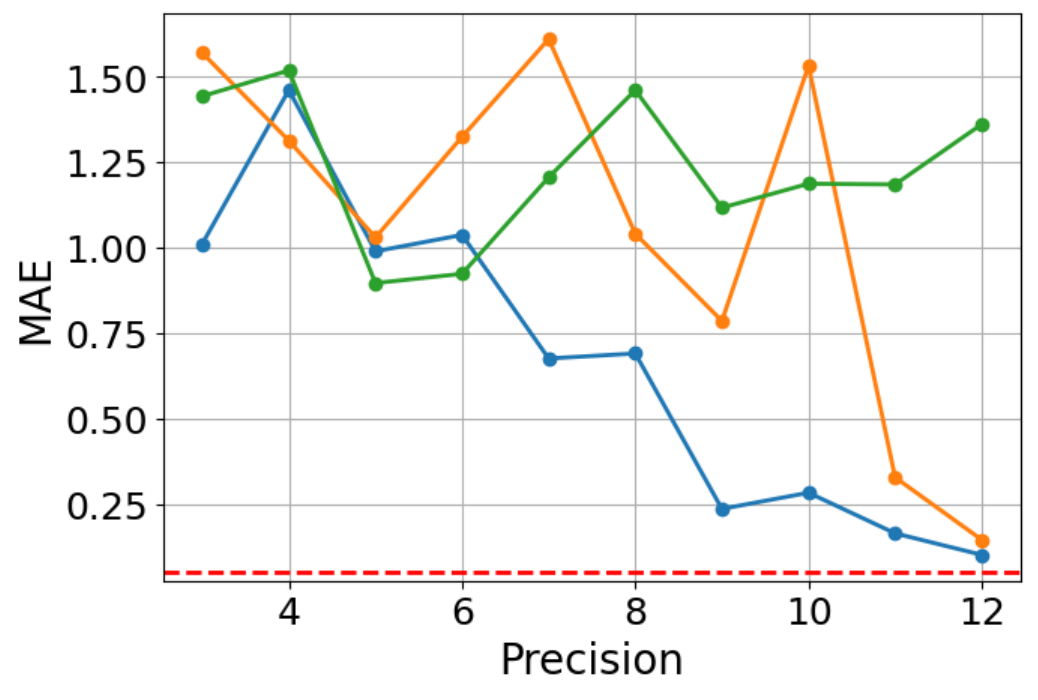}%
            \label{subfig:fusion-fkeras}%
        }
        \caption{Evaluation of ECON-T models (top row) and Fusion models (bottom row) robustness under different input or weight perturbations. Each subplot represents performance benchmarks on specific scenarios: (a and e) clean data, (b and f) perturbed data with Gaussian noise, (c and g) perturbed data with salt-and-pepper noise, and (d and h) flipping the five most sensitive bits. The models are trained with three regularization methods: Baseline (no regularization), Jacobian regularization, and orthogonal regularization.}
        \label{fig:benchmarks}
\end{figure*}

\textbf{Input Perturbations.}
The robustness of the ECON-T model against input perturbations is critical to ensuring the reliability of processed data. Regularization techniques are promising candidates to improve model robustness; however, they may degrade performance on clean data, as shown in Figures \ref{subfig:econ-clean} and \ref{subfig:fusion-clean}. While Jacobian regularization introduces negligible performance degradation, the degradation caused by orthogonal regularization may be unacceptable depending on the use case. The regularization weight $\delta$ in the optimization loss (see Eqs.~\eqref{eqn:reg-jac} and~\eqref{eqn:reg-ort}) can be adjusted to manage the trade-off between clean-data performance and model robustness. Fine-tuning this trade-off often requires several iterations to test various noise types and magnitudes to guarantee reliability. The loss landscape metrics previously introduced provide valuable insights that align with the performance results in Figure \ref{fig:benchmarks}. Noise significantly affects the performance of baseline models, as shown in Figures \ref{subfig:econ-gaussian} and \ref{subfig:fusion-gaussian} for Gaussian noise, and Figures \ref{subfig:econ-s_p} and \ref{subfig:fusion-s_p} for salt-and-pepper noise. However, mitigation techniques are not always effective in increasing robustness, and the reasons can be identified by looking at the loss landscape. Indeed, by matching the performance of the ECON-T model with the analysis of the loss landscape of discussed above, we can note that improvements in robustness provided by orthogonal regularization (Figs.~\ref{subfig:econ-gaussian}-\ref{subfig:econ-s_p}) correspond to a flatter and smoother landscape (Fig.~\ref{subfig:econ-mc}). In contrast, Jacobian regularization is less effective in mitigating noise, which correlates with poorer minimum connectivity and steeper slopes at convergence points in the landscape (Fig.~\ref{subfig:econ-mc}). For the Fusion model, the effectiveness of mitigation techniques is evident for high precisions only. Although baseline models converge to flatter minima in these cases, they tend to produce highly divergent representations (Fig.~\ref{subfig:fusion-cka}), which indicate obstacles in the optimization process that obstruct the reach of lower minima. Interestingly, both models exhibit higher robustness to input perturbations under low-bit quantization, a behavior that was also observed in previous studies~\cite{defensive}.

\textbf{Weight Perturbations.} Given the size of the analyzed models, the impact of weight perturbations (bit flips) on their performance is highly destructive. Nevertheless, the results of Figures \ref{subfig:econ-fkeras} and \ref{subfig:fusion-fkeras} confirm the observations made above also for this issue. Specifically, ECON-T models fine-tuned with low-bit configurations demonstrate greater robustness to weight perturbations, even with fewer bits per parameter. This behavior strongly correlates with the Hessian trace (Fig.~\ref{subfig:econ-trace}) and mode connectivity (Fig.~\ref{subfig:econ-mc}) analyses. Models with fewer bits per parameter tend to have a lower Hessian trace, i.e., indicating convergence to flatter minima, and reduced barriers between minima. For most configurations, except for extreme low-bit settings (e.g., 3 or 4 bits) where quantization and performance degradation are more pronounced, these phenomena are strongly correlated. Additionally, models trained with orthogonal regularization exhibit greater robustness to weight perturbations across most quantization configurations. This result aligns with their favorable Hessian trace and mode connectivity characteristics. In contrast, the Fusion model exhibits a different behavior. Performance degradation decreases as precision increases, consistently with the trends observed in Figures \ref{subfig:fusion-trace} and \ref{subfig:fusion-mc}. Low-precision Fusion models show more barriers between minima, and the baseline version of the model has the lowest Hessian trace compared to the regularized versions, explaining the results in Figure \ref{subfig:fusion-fkeras}.

\section{Conclusion}
\label{sec:conclusion}
We proposed a method to conduct a comprehensive empirical analysis of the loss landscape of machine learning models and applied the method to two representative, yet diverse models for scientific applications. These models require quantization to be deployed and are subject to noise and bit flips in the model parameters. Two regularization techniques were considered to mitigate noise and bit flips, complementing the intrinsic regularization provided by quantization.

Most interestingly, contrary to what one may expect, we found that increasing quantization precision does not always provide benefits in terms of robustness to noise and bit flips. Furthermore, we found that different models may benefit from different regularization techniques.

Our method allows efficient exploration of the trade-offs between robustness and performance without calling for tedious and time-consuming training campaigns for design space exploration. It does so without assuming prior knowledge of the perturbations. Automatic Pareto optimization of model configurations can hence be enabled by building on our method and represents our prominent future work.
%



\section*{Acknowledgements}
We thank Ryan Forelli for his help in this work. We also acknowledge the Fast Machine Learning collective as an open community of multi-domain experts and collaborators. This community was important for the development of this project.

TB and AB were supported by SERICS (PE00000014) under the MUR National Recovery and Resilience Plan funded by the European Union - NextGenerationEU, and OPERAND PRIN-PNRR project.

JC and NT were supported by the U.S. Department of Energy (DOE), Office of Science,
Office of Advanced Scientific Computing Research under the “Real-time Data Reduction
Codesign at the Extreme Edge for Science” Project (DE-FOA-0002501).

OW was supported by the NSF Graduate Research Fellowship Program under Grant No. DGE-2038238. 
Any opinions, findings, and conclusions or recommendations expressed in this material are those of the author(s) and do not necessarily reflect the views of the NSF.

CG was supported by the U.S. Department of Energy (DOE), Office of Science, Advanced Scientific Computing Research (ASCR) program under Contract Number DE-AC02-05CH11231 to Lawrence Berkeley National Laboratory (``Visualizing High-dimensional Functions in Scientific Machine Learning'').


\section*{Impact Statement}
This work contributes to the advancement of ML research by being the first reliability analysis of ML models for scientific sensing applications based on the loss landscape, and also copes with their training strategy to make them more reliable against unknown corruptions. 
Using loss landscape analysis, we provide actionable a-priori insights that minimize the need for brute-force testing, which typically requires extensive iterative experimentation. 
This approach can significantly reduce training resources and provide robustness against out-of-training-distribution corruptions. Furthermore, our analysis is grounded in models that are actively used in scientific experiment -- improving control of complex systems and  accelerating scientific research and, thus, potential discoveries. By enhancing the efficiency and reliability of ML models in these critical domains, our methodology aligns with the broader goals of sustainable and impactful AI research and development.





\bibliography{main}
\bibliographystyle{icml2025}

\newpage
\appendix
\onecolumn
\section{Example of mode connectivity computation}
\label{sec:mc_ex}
In this section, we provide a more detailed example of how mode connectivity is computed. Please refer to Section~\ref{sec:mc} of the paper.

\begin{figure}[ht!]
        \centering
        \includegraphics[width=0.6\linewidth]{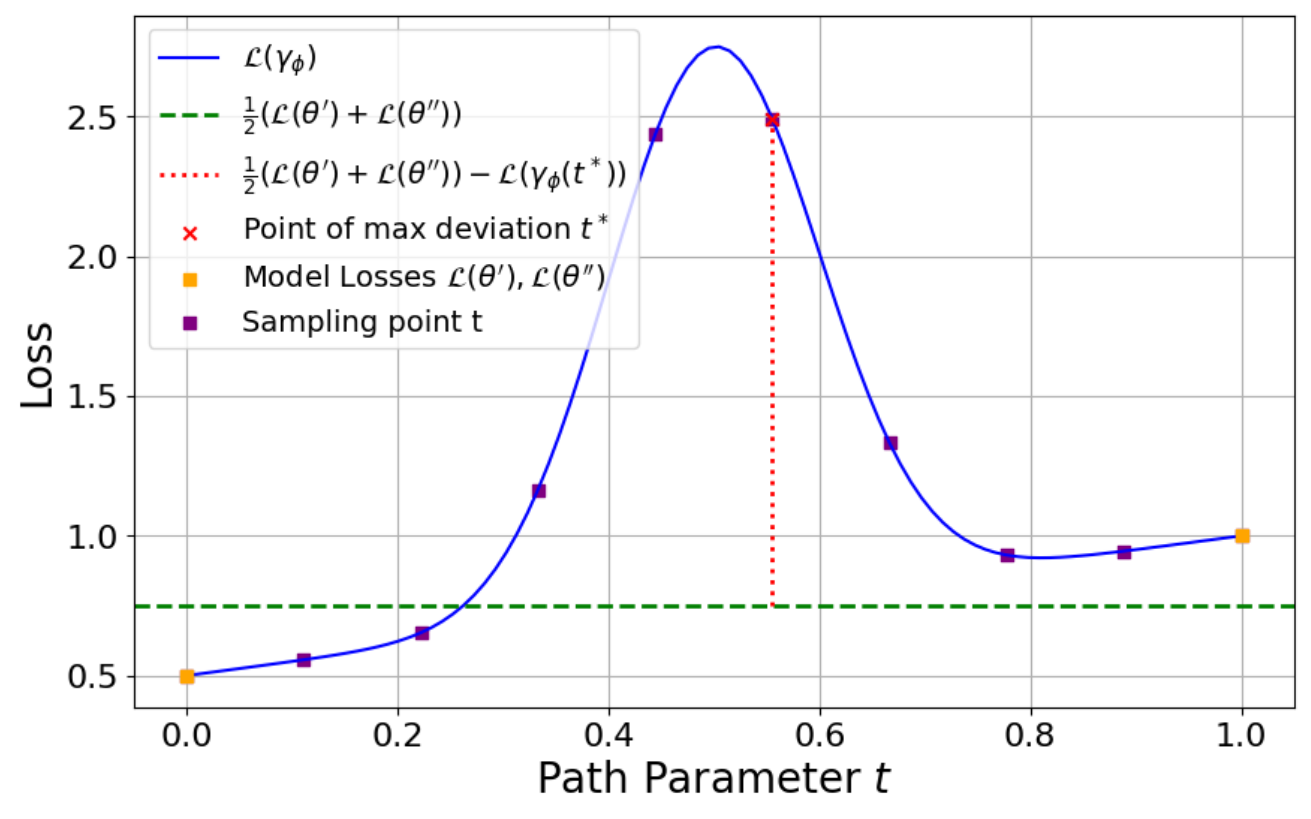}%
        \caption{Example of mode connectivity computation. The blue line represent the loss computed along the Bezier curve $\gamma_\phi$. We sample $m$ points along this curve, and then we look for the point which is maximizing the deviation from the average loss ($t^*$ in Section~\ref{sec:mc}) between the two extreme model parameters $\theta'$ and $\theta''$.}
        \label{fig:mc-example}
\end{figure}

Once the Bezier curve between two models, with parameters $\theta'$ and $\theta''$, is defined, we can then sample an arbitrary number of points along this curve by picking a value $t \in [0, 1]$. The extreme cases, $t = 0$ and $t = 1$, correspond to $\theta'$ and $\theta''$, respectively. Each intermediate point $t_i$ for $i = 1, ..., m-2$ serves as input to the Bezier curve, yielding a possible configuration of model parameters. These parameters can be used to evaluate the loss along the Bezier curve, as illustrated by the blue line in Figure \ref{fig:mc-example}).

\section{Ablation study on metrics}
\label{sec:metric_abl}
In this section, we first compare the loss landscapes visualization method proposed by~\cite{visualizing} with our approach, and then investigate different configurations when measuring the CKA similarity and the mode connectivity.
\begin{figure}[ht!]
        \centering
        \subfloat[]{%
            \includegraphics[width=0.31\linewidth]{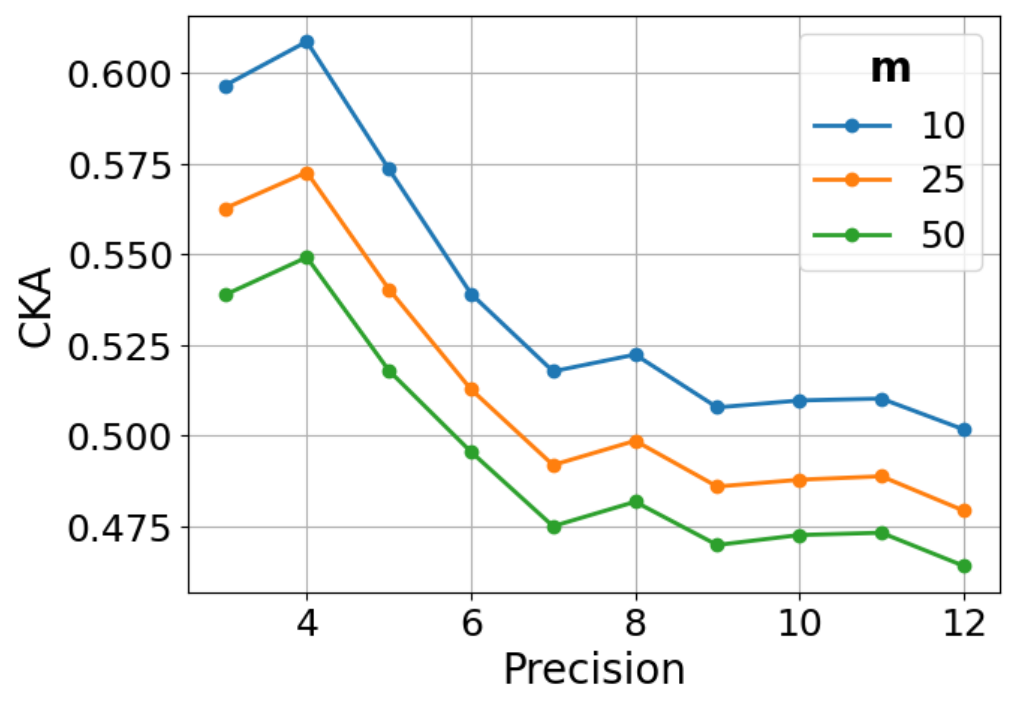}%
            \label{subfig:abl-cka-m}%
        }%
        \subfloat[]{%
            \includegraphics[width=0.32\linewidth]{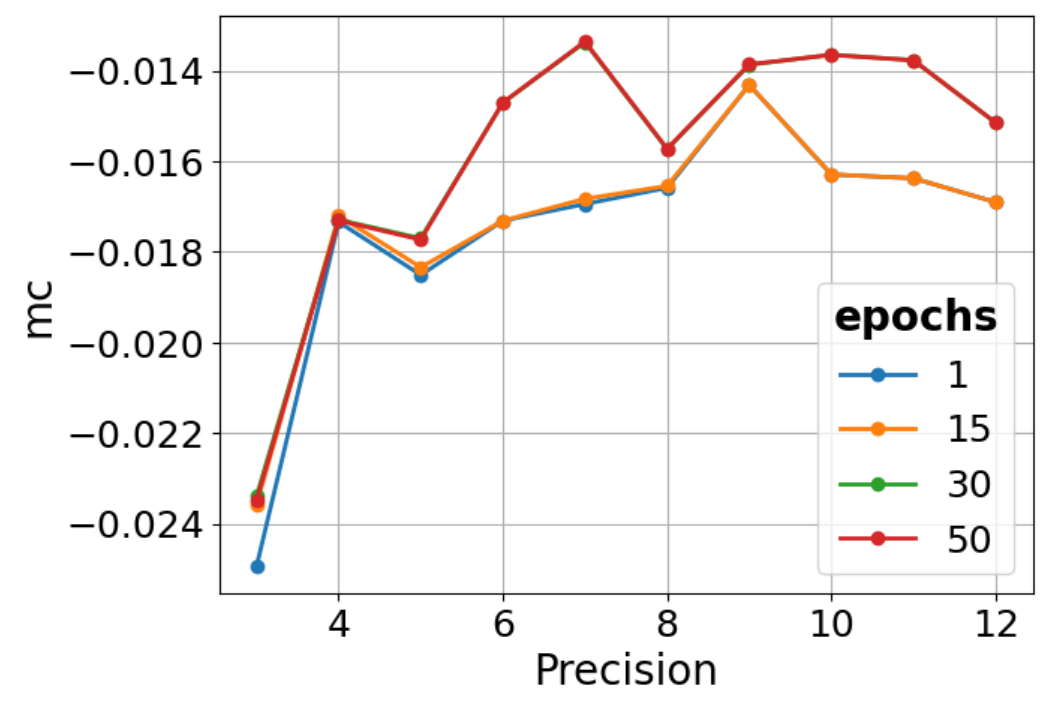}%
            \label{subfig:abl-mc-epochs}%
        }%
        \subfloat[]{%
            \includegraphics[width=0.298\linewidth]{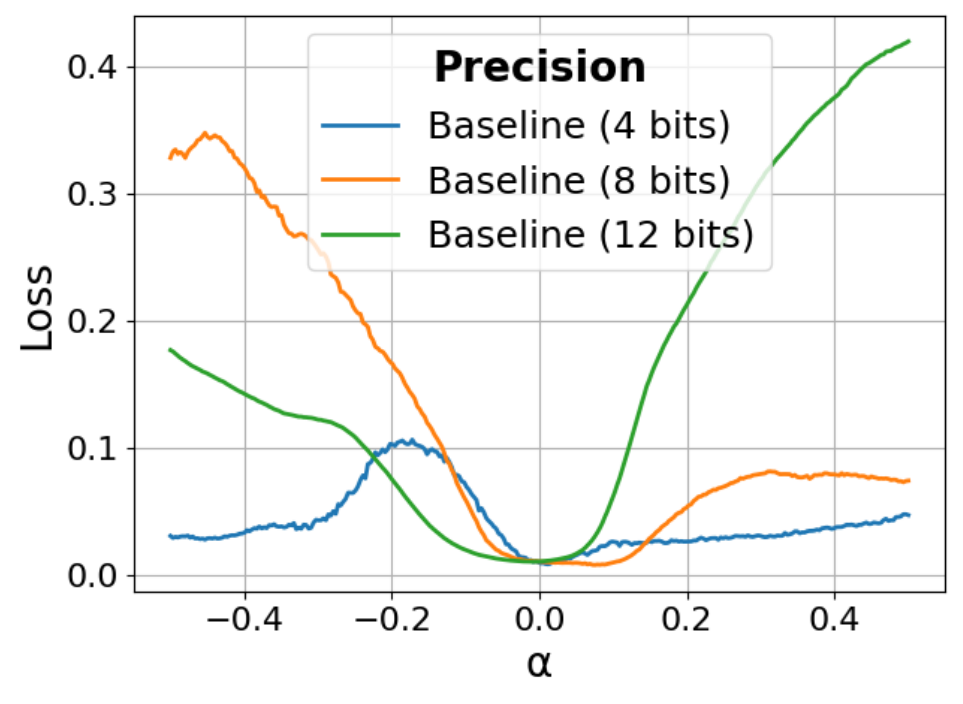}%
            \label{subfig:abl-vis-hessian}%
        }\\%
        \centering
        \subfloat[]{%
            \includegraphics[width=0.3\linewidth]{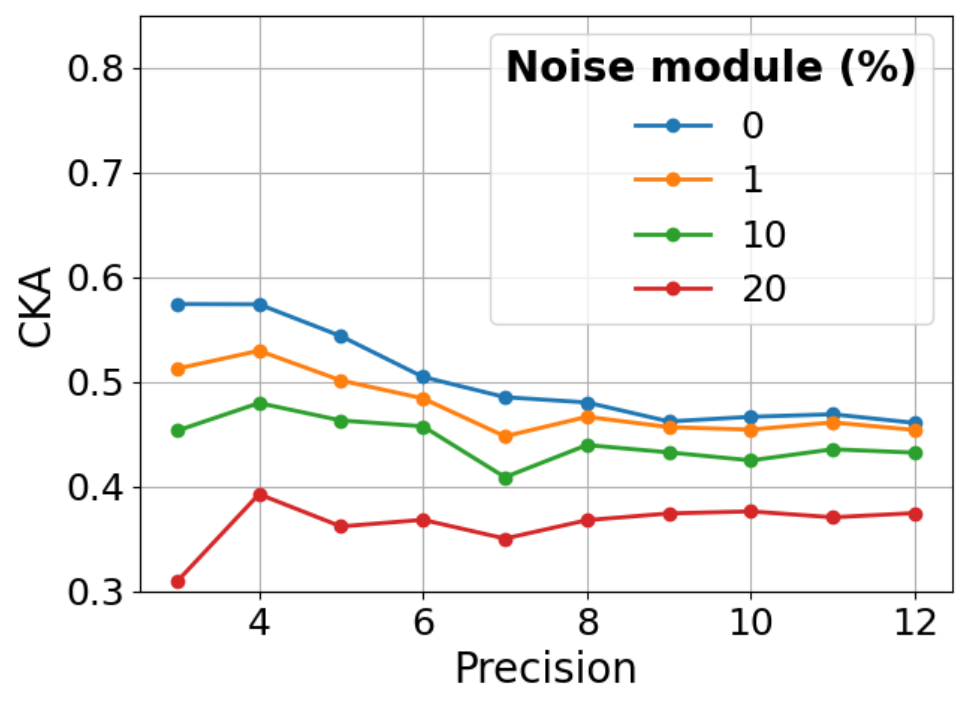}%
            \label{subfig:abl-cka-noise}%
        }%
        \subfloat[]{%
            \includegraphics[width=0.325\linewidth]{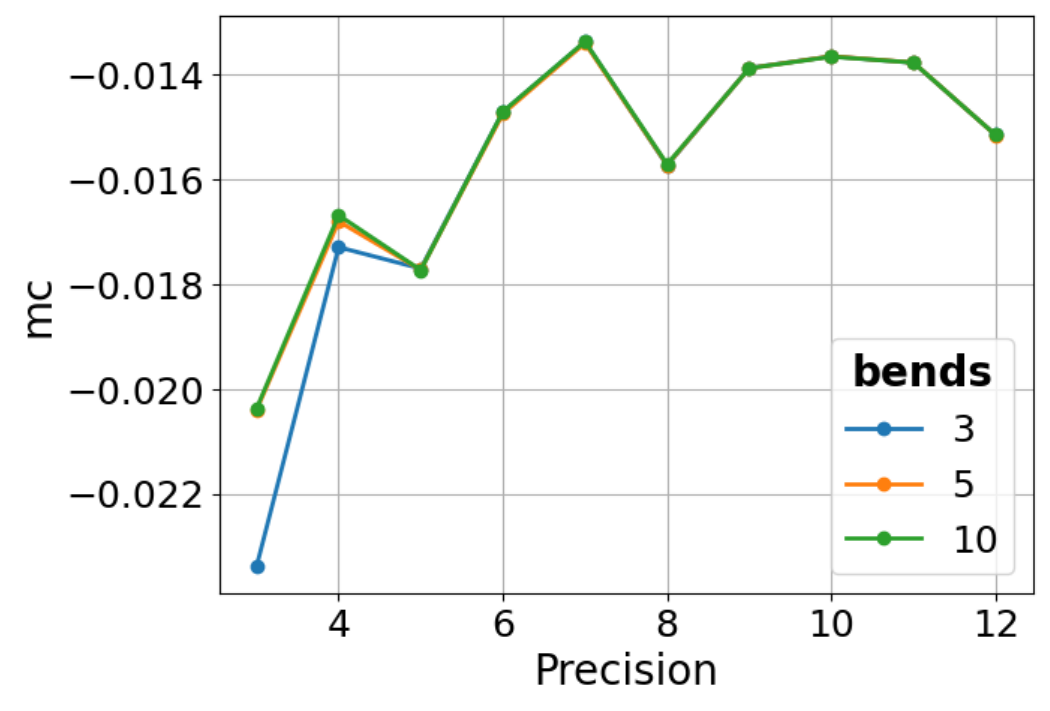}%
            \label{subfig:abl-mc-bends}%
        }%
        \subfloat[]{%
            \includegraphics[width=0.315\linewidth]{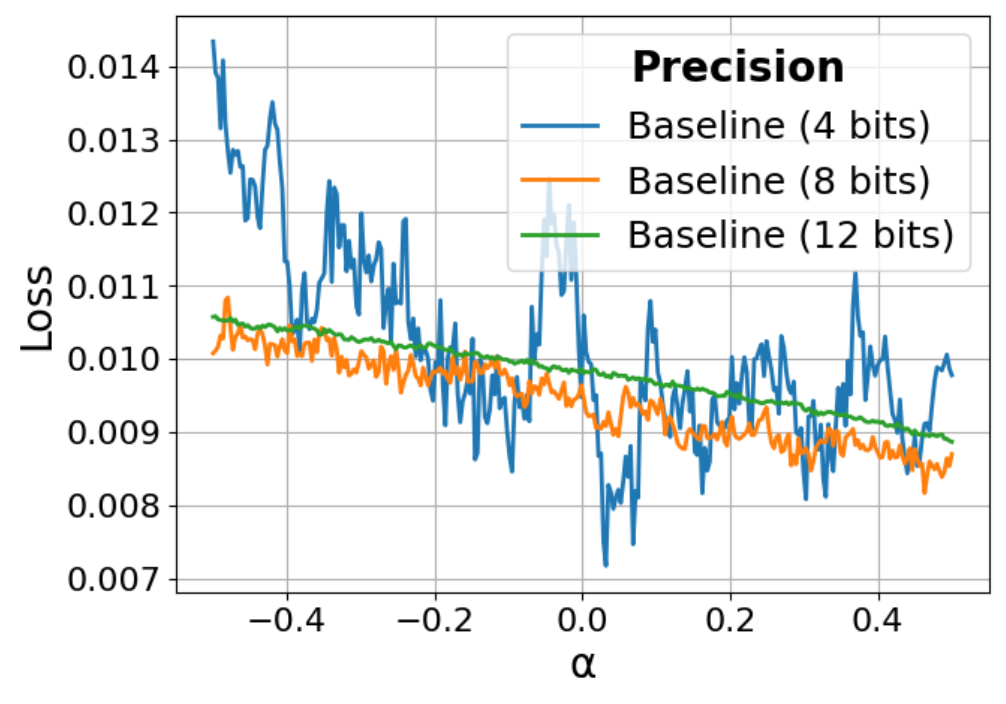}%
            \label{subfig:abl-vis-random}%
        }
        \caption{Ablation studies of the loss landscape metrics. Subplots show: (a) and (d) \textbf{CKA similarity} of the ECON-T model where we respectively explore the impact of changing the number of concatenated outputs $m$, and the noise intensity; (b) and (e) \textbf{mode connectivity} of the Fusion model, comparing results obtained tuning the number of training epochs (b) and the number of bends (e); (c) and (f) \textbf{loss landscape visualization} methods, where the directions are computed respectively with top eigenvector of model parameters and a random direction.}
        \label{fig:abl-benchmarks}
\end{figure}

\subsection{Ablation study on loss landscape visualization}
\label{sec:abl-vis}
We start by showing why visualizing the loss landscape using the top eigenvector as the perturbation direction (Figure \ref{subfig:abl-vis-hessian}), our novel proposal that differs from the approach in~\cite{visualizing},  provides more informative insights compared to using random directions (Figure \ref{subfig:abl-vis-random}). Although both plots depict the loss landscape of the same Fusion models, they exhibit notable differences.

First, by examining the y-axis, we observe that the scales of the two plots are completely different. When using random directions—computed within the same range and with the same resolution (i.e., number of steps)—the loss variation is negligible. As a result, this approach fails to provide meaningful insights into suboptimal minima and their sharpness.

The only notable feature in Figure \ref{subfig:abl-vis-random} is the jagged shape introduced by low-bit quantization. However, this effect can be disregarded, as we have no guarantee that the model will follow this direction during training. In contrast, when adopting the top eigenvector as the perturbation direction, we ensure that the visualization captures the model’s behavior along the direction of maximum curvature. This is particularly relevant because most training optimizers, such as SGD, tend to explore this direction, making it a more reliable choice for loss landscape analysis.

\subsection{Ablation study on measuring CKA similarity}
\label{sec:abl-cka}
In this subsection, we provide a detailed analysis of how the number of concatenated outputs and the intensity of input perturbations impact the CKA results. As shown in Figure \ref{subfig:abl-cka-m}, increasing the number of concatenated outputs leads to lower CKA similarity. This outcome is expected, as increasing the dimensionality of the matrices being compared raises the likelihood of differences between them. However, we observe that the overall patterns remain consistent, demonstrating the robustness of the metric across different configurations. In this work we used $m=10$ to save computation time. Additionally, it is often preferable to conduct this analysis on perturbed data distributions, which can be achieved by adding uniformly distributed noise, such as Gaussian noise, to the inputs. This approach is particularly useful when models are trained to achieve near-zero training loss, as it enhances the informativeness of the metric. However, Figure \ref{subfig:abl-cka-noise} indicates that this is not necessary for the ECON-T model. Even without noise injection, we can extract meaningful insights regarding the CKA similarity of the models.

\subsection{Ablation study on measuring mode connectivity}
\label{sec:mc-abl}
In this subsection, we analyze the optimal configuration for generating the mode connectivity plots shown in Figures \ref{subfig:econ-mc} and \ref{subfig:fusion-mc}. The first key parameter to consider is the number of training epochs for the models used to construct the Bezier curve. This hyperparameter is particularly important because the characteristic curved shape of the Bezier curve emerges only when the intermediate models begin converging to their respective minima. Otherwise, the sampled models will lie along the linear interpolation between the two models, leading to a coarse approximation.

Figure \ref{subfig:abl-mc-epochs} highlights this effect. Specifically, training the intermediate models for 1 and 15 epochs produces similar results, whereas training for 30 and 50 epochs leads to significantly different outcomes. Since the curves for 30 and 50 epochs are nearly identical, we opted for 30 epochs to reduce computational cost in subsequent experiments.

Another crucial hyperparameter in mode connectivity analysis is the number of bends $k+1$, which determines the complexity of the Bezier curve. Figure \ref{subfig:abl-mc-bends} shows that increasing the number of bends has little impact, likely due to the relatively low number of parameters in the model under analysis. This results in a non-trivial loss landscape morphology that does not require highly parameterized Bezier curves. Therefore, we chose to use three bends to balance accuracy and computational efficiency.

\section{Ablation study on benchmarks and mitigation techniques}
\label{sec:loss_abl}
In this section, we study different configurations of regularization methods and benchmarks.

\subsection{Ablation study on benchmarks}
\label{sec:bench_abl}
\begin{figure}[ht!]
        \centering
        \subfloat[]{%
            \includegraphics[width=0.26\linewidth]{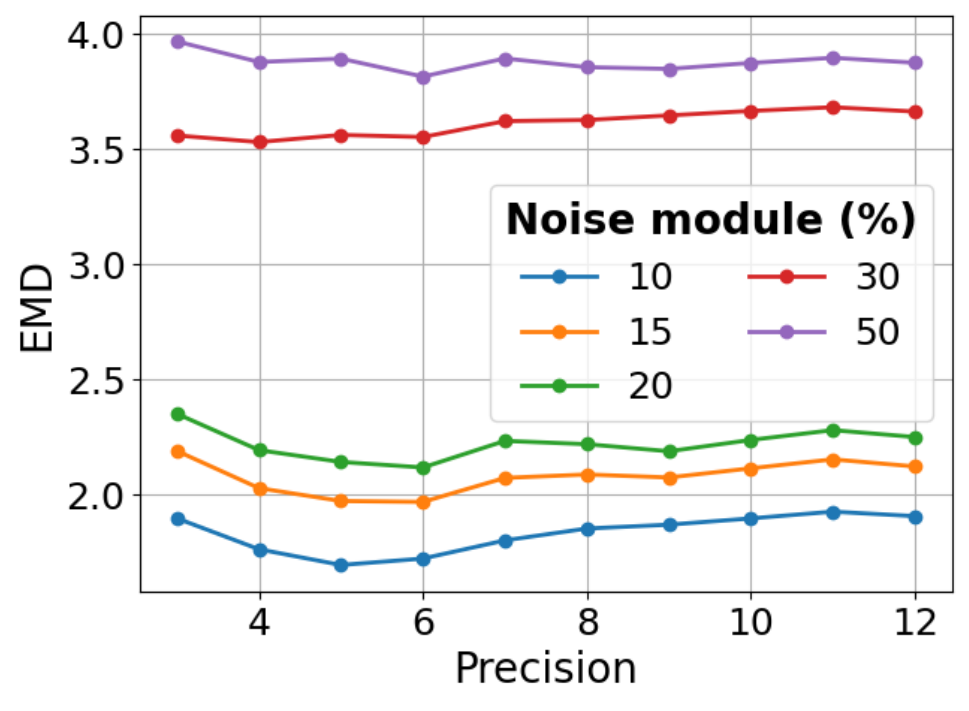}%
            \label{subfig:abl-gaussian}%
        }%
        \subfloat[]{%
            \includegraphics[width=0.26\linewidth]{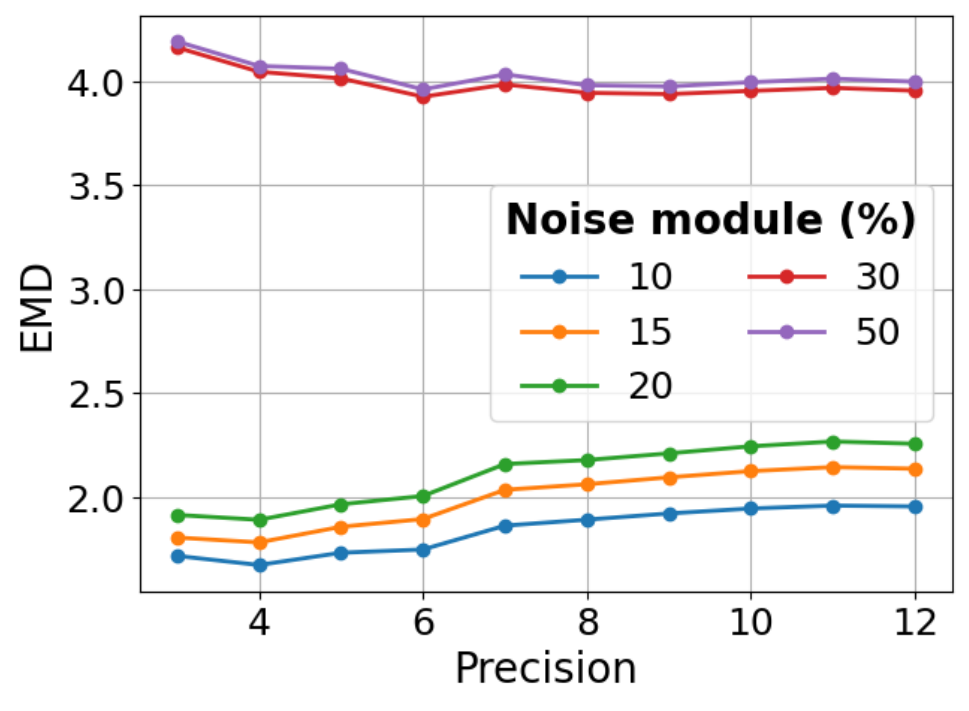}%
            \label{subfig:abl-sp}%
        }%
        \subfloat[]{%
            \includegraphics[width=0.25\linewidth]{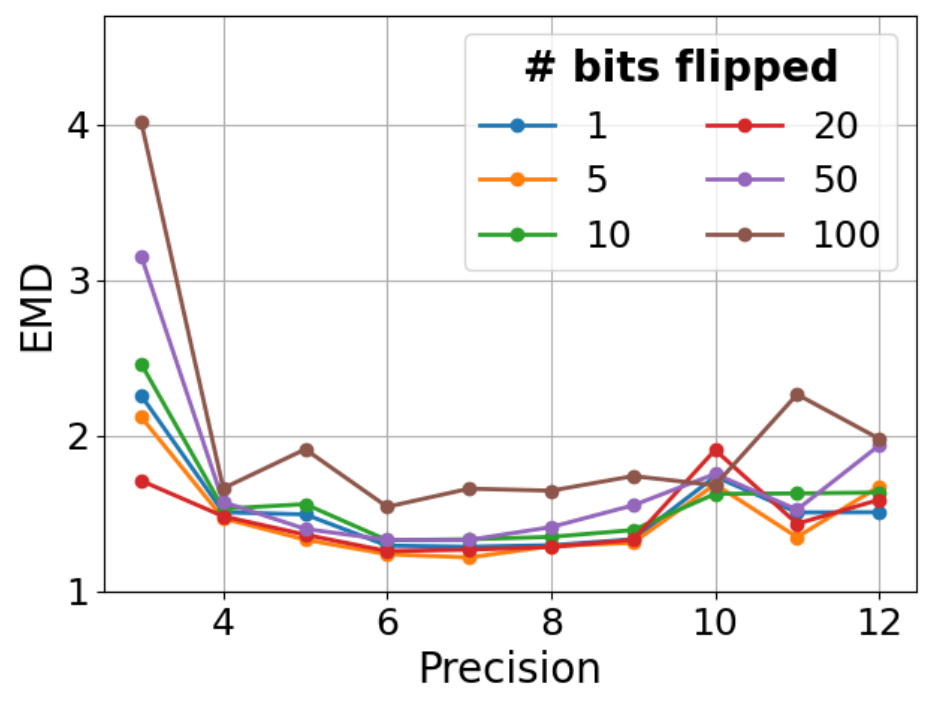}%
            \label{subfig:abl-random}%
        }%
        \subfloat[]{%
            \includegraphics[width=0.25\linewidth]{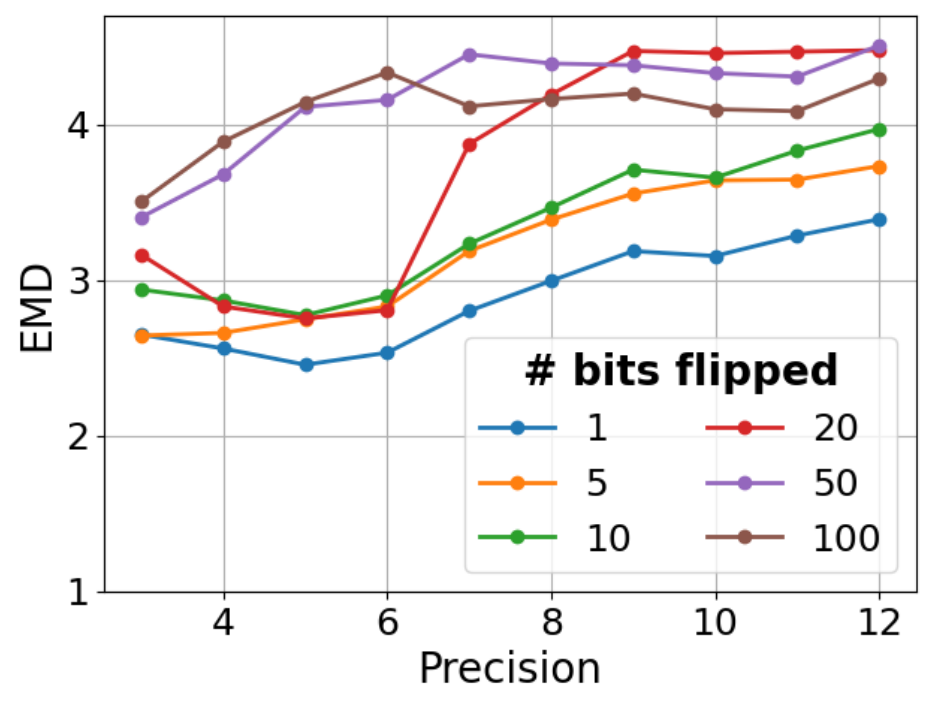}%
            \label{subfig:abl-fkeras}%
        }%
        \caption{Evaluation of the ECON-T model under different stress conditions: (a) and (b) shows respectively the performances of the model where the input is corrupted with Gaussian and salt-and-pepper noise, focusing the attention on the models reliability respect to different noise intensity; (c) and (d), instead, compare the degradation of performances of the models where the parameters are perturbed by different numbers of bit errors,the bits to be flipped are picked randomly in (c) and adopting FKeras methodology in (d).}
        \label{fig:abl-benchmarks}
\end{figure}

In this subsection, we provide a comprehensive evaluation of the performance of the baseline version of the ECON-T model under varying noise magnitudes (Figures \ref{subfig:abl-gaussian} and \ref{subfig:abl-sp}) and bit error rates (Figures \ref{subfig:abl-random} and \ref{subfig:abl-fkeras}), analogous to the analysis conducted in Figure \ref{fig:benchmarks}.

Regarding input perturbations, both noise types exhibit similar behavior: for noise intensities below 20\%, performance degradation is observed, but the model still follows the same overall pattern. However, at higher noise intensities, the destructive effects become irrecoverable and unpredictable.

For parameter perturbations, we validate the effectiveness of the FKeras approach as a benchmark for the worst-case scenario by comparing Figures \ref{subfig:abl-random} and \ref{subfig:abl-fkeras}. Notably, not all bits contribute equally to model performance, as evidenced by the fact that flipping the most sensitive bit, as identified by FKeras, leads to significantly more destructive effects than randomly flipping 100 bits.

\begin{figure}[th!]
        \centering
        \subfloat[]{%
            \includegraphics[width=0.26\linewidth]{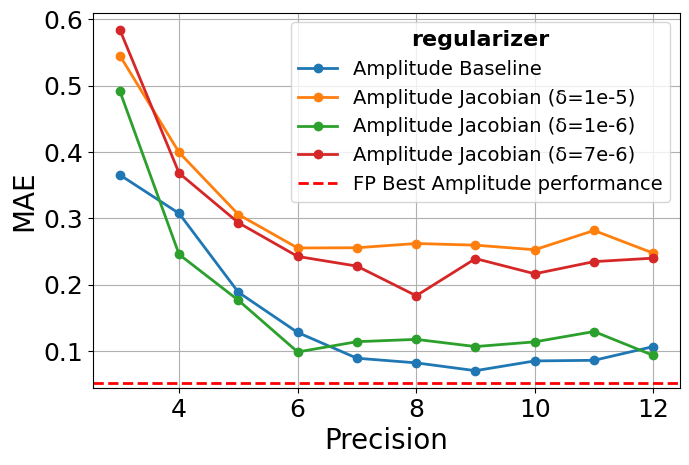}%
            \label{subfig:abl-jreg-clean}%
        }%
        \subfloat[]{%
            \includegraphics[width=0.26\linewidth]{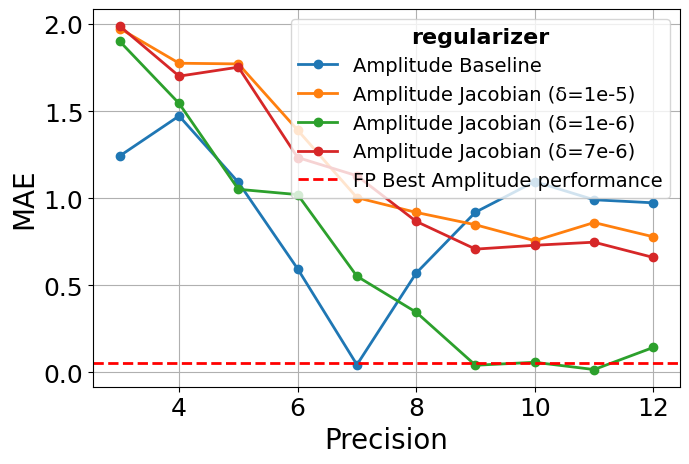}%
            \label{subfig:abl-jreg-noise}%
        }%
        \subfloat[]{%
            \includegraphics[width=0.26\linewidth]{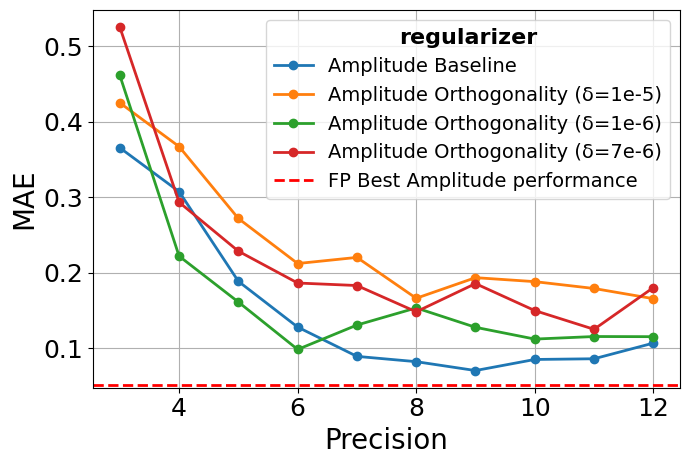}%
            \label{subfig:abl-lip-clean}%
        }%
        \subfloat[]{%
            \includegraphics[width=0.26\linewidth]{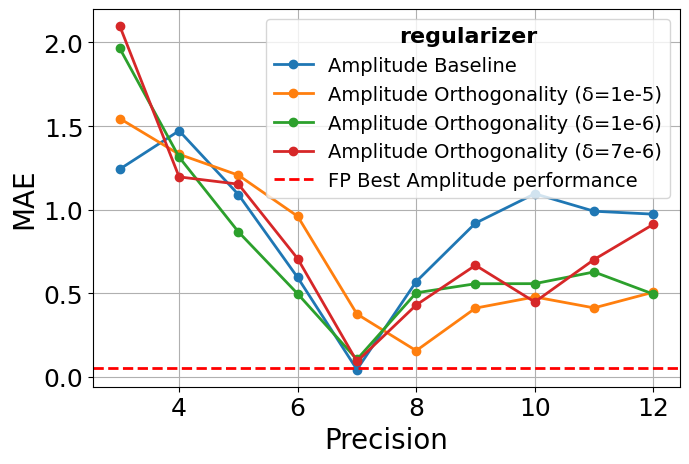}%
            \label{subfig:abl-lip-noise}%
        }%
        \caption{Evaluation of the Fusion model fine-tuned with different values of the coefficient $\delta$ for the regularization part of the loss: (a) and (b) shows respectively the performances of the model on clean and perturbed input (10\% Gaussian noise), changing the $\delta$ of the Jacobian regularization; (c) and (d), instead, shows respectively the performances of the model on clean and perturbed input (10\% Gaussian noise), changing the $\delta$ of the orthogonal regularization.}
        \label{fig:abl-benchmarks}
\end{figure}

\subsection{Ablation study on mitigation techniques tuning}
\label{sec:reg_abl}

In this subsection, we provide a comprehensive evaluation of the impact of the regularization techniques proposed in this work, comparing their performance on both clean data (Figures \ref{subfig:abl-jreg-clean} and \ref{subfig:abl-lip-clean}) and perturbed data (Figures \ref{subfig:abl-jreg-noise} and \ref{subfig:abl-lip-noise}). The model under analysis is the Fusion model, trained with different regularization coefficients $\delta$ for each regularizer.

The coefficient $\delta$ controls the weight of the regularization term in the loss function, determining the trade-off between performance on clean and perturbed data. Intuitively, a lower $\delta$ results in minimal performance degradation on clean data while offering limited robustness improvement. Conversely, a higher $\delta$ may significantly enhance robustness but at the cost of greater performance degradation on clean data.

In this study, we tested different values of $\delta$ and selected the one that provided the best trade-off ($\delta = 10^{-6}$ in this case). While more sophisticated tuning methods could be explored, they are beyond the scope of this work.


\end{document}